\documentclass[preprint,12pt]{elsarticle}

\usepackage{lscape}

\usepackage[detect-all]{siunitx}
\DeclareSIUnit\pixel{px}

\usepackage{pgfplots}
\usepackage{hyperref}
\usepackage{todonotes}
\usepackage{setspace}
\usepackage{textcomp}

\usepackage{geometry} 
\usepackage{epstopdf}
\usepackage{float}
\usepackage{multirow}
\usepackage{tabularx}
\usepackage{booktabs}

\usepackage{graphicx}
\usepackage{amsmath,epsfig}
\usepackage{cases}

\usepackage{subfigure}
\usepackage{array}

\usepackage{url}
\makeatletter
\g@addto@macro{\UrlBreaks}{\UrlOrds}
\makeatother

\usepackage{caption}
\usepackage{makecell}
\usepackage{cellspace}
\usepackage{graphicx}
\usepackage{caption}
\DeclareCaptionFormat{cont}{#1 (cont.)#2#3\par}

\usepackage{epstopdf}
\usepackage{mathtools} 
\usepackage{comment}
\usepackage[normalem]{ulem}
\usepackage{pgfplotstable}

\AppendGraphicsExtensions{.tif}

\newcolumntype{v}[1]{>{\raggedright\hspace{0pt}}p{#1}}

\usepackage{tikz}
\usepackage{xcolor}
\usepackage{colortbl}
\definecolor{tumblue}{RGB}{0,101,189}
\definecolor{tumblack}{rgb}{0, 0, 0}
\definecolor{tumwhite}{rgb}{1, 1, 1}
\definecolor{tumbluedark}{RGB}{0,82,147}
\definecolor{tumbluelight}{RGB}{152,198,234}
\definecolor{tumbluemedium}{RGB}{100,160,200}
\definecolor{tumgray}{gray}{0.6}
\definecolor{tumgray1}{gray}{0.2}
\definecolor{tumgray2}{gray}{0.4}
\definecolor{tumgray4}{gray}{0.6}

\definecolor{tumivory}{RGB}{218,215,203}
\definecolor{tumgreen}{RGB}{162,173,0}
\definecolor{tumorange}{RGB}{227,114,34}
\colorlet{tumgraylight}{tumgray!25!white}
\colorlet{tumgraydark}{tumgray!50!black}
\definecolor{tumdiagramaubergine}{RGB}{105,8,90}
\definecolor{tumdiagramnavyblue}{RGB}{15,27,95}
\definecolor{tumdiagramturquoise}{RGB}{0,119,138}
\definecolor{tumdiagramgreen}{RGB}{0,124,48}
\definecolor{tumdiagramlimegreen}{RGB}{103,154,29}
\definecolor{tumdiagramyellow}{RGB}{255,220,0}
\definecolor{tumdiagramsand}{RGB}{249,186,0}
\definecolor{tumdiagramredorange}{RGB}{214,76,19}
\definecolor{tumdiagramred}{RGB}{196,72,27}
\definecolor{tumdiagramdarkred}{RGB}{156,13,22}
\definecolor{compHR}{RGB}{133 ,   6,       10}
\definecolor{compMR}{RGB}{199,   14,        21}
\definecolor{compLR}{RGB}{243,   19,        29}
\definecolor{scatTree}{RGB}{21,   255,     21}
\definecolor{openHR}{RGB}{183 , 78 , 18}
\definecolor{openMR}{RGB}{244, 104 , 29}
\definecolor{openLR}{RGB}{ 247, 154 , 90}
\definecolor{lowPlant}{RGB}{204,   255,    102}

\definecolor{b3}{RGB}{224, 243, 219}
\definecolor{gugs}{RGB}{0, 0, 255}
\definecolor{gu}{RGB}{0, 255, 0}
\definecolor{gsp}{RGB}{255, 0, 0}
\definecolor{gh}{RGB}{119,	172,	48}
\definecolor{gup}{RGB}{77,	190,	238}
\definecolor{p}{RGB}{156, 80, 242}
\makeatletter
\newenvironment{customlegend}[1][]{%
\begingroup
    \let\addlegendimage=\pgfplots@addlegendimage
    \let\addlegendentry=\pgfplots@addlegendentry

    \pgfplots@init@cleared@structures
    \pgfplotsset{#1}%
}{
    \pgfplots@createlegend
    \endgroup
}
\makeatother

\journal{}

\begin{document}

\begin{frontmatter}

\title{
A framework for large-scale mapping of human settlement extent from Sentinel-2 images via fully convolutional neural networks}

\author[firstaddress]{Chunping Qiu}
\author[firstaddress]{Michael Schmitt}
\author[secondaryaddress]{Christian Geiß}
\author[thirdaryaddress]{Tzu-Hsin Karen Chen}
\author[firstaddress,secondaryaddress]{Xiao Xiang Zhu\corref{correspondingauthor}}
\cortext[correspondingauthor]{xiaoxiang.zhu@dlr.de}

\address[firstaddress]{Signal Processing in Earth Observation (SiPEO), Technical University of Munich (TUM), Arcisstr. 21, 80333 Munich, Germany}
\address[secondaryaddress]{Remote Sensing Technology Institute (IMF), German Aerospace Center (DLR),  Oberpfaffenhofen, 82234 Wessling, Germany}
\address[thirdaryaddress]{Department of Environmental Science, Aarhus University, Frederiksborgvej 399, DK-4000 Roskilde, Denmark}

\begin{abstract}\textcolor{blue}{This article was submitted to ISPRS Journal of Photogrammetry and Remote Sensing.}

Human settlement extent (HSE) information is a valuable indicator of world-wide urbanization as well as the resulting human pressure on the natural environment. Therefore, mapping HSE is critical for various environmental issues at local, regional, and even global scales. This paper presents a \textcolor{black}{deep-learning-based framework} to automatically map HSE from multi-spectral Sentinel-2 data {using regionally available geo-products as training labels}. \textcolor{black}{A straightforward, simple, yet effective fully convolutional network-based architecture, Sen2HSE, is implemented as an example for semantic segmentation within the framework.} The framework is validated against both manually labelled checking points distributed evenly over the test areas, and the OpenStreetMap building layer. The HSE mapping results were extensively compared to several baseline products in order to thoroughly evaluate the effectiveness of the proposed HSE mapping framework. The HSE mapping power is consistently demonstrated {over 10 representative areas across the world}. \textcolor[rgb]{0,0,0}{We} also present one regional-scale and one country-wide HSE mapping example from our framework to show the potential for upscaling. The results of this study contribute to the generalization of the applicability of CNN-based approaches for {large-scale} urban mapping to cases where no up-to-date and accurate ground truth \textcolor{black}{is} available, as well as the subsequent monitor of global urbanization.



\end{abstract}
  

\end{frontmatter}

\section{Introduction}
\label{sec:intro}
Human settlement extent (HSE), which is characterized by buildings, roads, and other man-made structures, is an essential indicator of the human footprint on the Earth. Moreover, it is an expression of the {impact of} ongoing worldwide urbanization. According to \cite{united20182018}, 55\% of the world's population now lives in urban areas, a proportion that is expected to increase to 68\% by 2050. 
{To better understand drivers and interactions between urbanization and social and environmental processes, it is \textcolor{black}{thus} necessary to obtain accurate and up-to-date HSE data.}
Recent years have seen a proliferation of studies related to HSE mapping, among which remote sensing-based approaches have gained more and more attention due to their inherent ability to frequently and regularly observe the land surface on a global scale. With this unique property, several remote sensing-based global products related to HSE have become available. One, the Global Urban Footprint (GUF), was derived using TerraSAR-X as well as TanDEM-X Synthetic Aperture Radar (SAR) images \cite{esch2012tandem, esch2013urban}. Another, the Global Human Settlement (GHS) built-up grid, was derived from the Landsat as well as the Sentinel-1 image collections. GHS built-up grid is a product derived within the GHSL image analytics framework, which also utilizes remote sensing images from other missions such as SPOT-5 and 6 \cite{pesaresi2016operating, corbane2017big}. Still others, the GlobeLand30 land cover map and the Global Human Built-up And Settlement Extent (HBASE), were derived from the 30m resolution Landsat data \cite{chen2017analysis, Wang2017}. There are several other global land cover maps, such as {finer resolution observation and monitoring of global land cover with 30 m (FROM-GLC30) and 10 m (FROM-GLC10) resolution}, Global Land Cover 2000 (GLC2000) with 1 km resolution, and those derived from Moderate Resolution Imaging Spectrometer (MODIS) data with 500 m resolution, which are also produced using remote sensing image analysis \cite{gong2013finer, gong2019stable, bartholome2005glc2000, friedl2002global}. \textcolor{black}{It is difficult to compare these products directly as they each have slightly different foci. Generally, among these products, GUF outperforms the others \citep{marconcini2019outlining}, especially in rural areas where most of the products fail to detect impervious surfaces. GUF, however, is not feasible for frequent update as it was derived from the {relatively expensive} high resolution TerraSAR-X and TanDEM-X SAR images.}

Novel approaches for urban mapping explore cloud computing services like Google Earth Engine and the large amount of remote sensing data it offers \cite{patel2015multitemporal, goldblatt2018using, liu2018high}. In these examples, it is expected that the globally available multi-spectral Sentinel-2 data, with a 5-day temporal resolution and 10-meter spatial resolution, are going to play a key role in \textcolor{black}{more accurate} HSE mapping \textcolor{black}{at a large or even global scale, with the potential for frequent monitoring of global urbanization}. This is already being shown by some regional-scale studies, with similar applications on urban impervious surface mapping \cite{xu2018extraction} and land cover mapping \cite{gong2019stable, qiuRcnn}.

In the past, urban mapping approaches \textcolor{black}{typically} started \textcolor{black}{by} {extracting hand-crafted} features such as the normalized difference spectral vector (NDSV) and the gray-level co-occurrence matrix (GLCM), followed by feeding the extracted features into a traditional classifier such as Random Forests \cite{patel2015multitemporal, ban2015spaceborne, chini2018towards}, and ending with post-processing to remove potential mis-classifications. However, as a form of semantic segmentation task (or pixel level labeling), HSE mapping can theoretically be carried out through deep learning-based approaches, because plenty of neural network architectures have been proposed and shown to be powerful for semantic segmentation tasks. For example, SegNet, U-Net, the deconvolution network\textcolor{black}{,} as well as other improved variants based on multi-scale context fusion, attention mechanisms, and recurrent neural networks, were all proposed after fully convolutional networks (FCNs) were introduced in 2015 \cite{long2015fully,noh2015learning,badrinarayanan2017segnet,ronneberger2015u, badrinarayanan2017segnet}. The fundamental advantage of all these deep neural networks is their ability for enhanced feature representation  and pixel-level recognition. {Examples where convolutional neural networks (CNN) and, in particular, FCNs are used for remote sensing image classification or segmentation include \cite{paisitkriangkrai2016semantic, maggiori2016fully, langkvist2016classification, maggiori2016convolutional, fu2017classification, volpi2016dense, russwurm2018multi, zhang2019joint, zhong2019deep, hu2019mapping, lang2019country, he2018detecting}}. 
\textcolor{black}{Apart from the works focusing on very high resolution satellite or aerial imagery (i.e., with a ground sampling distance equal to or even less than 1 m), data of lower spatial resolution is also being studied, since the images of lower resolutions such as globally openly available Sentinel-2 imagery remain the key candidates for large-scale mapping \citep{helber2019eurosat, sumbul2019bigearthnet}.} 

\textcolor{black}{Good performance, however, is not guaranteed when directly employing} these existing approaches for large-scale HSE mapping from Sentinel-2 images. \textcolor{black}{There are three reasons for this, each with possible solutions}. First, getting sufficient reliable pixel-wise ground truth data, a major prerequisite for deep learning-based approaches, is more challenging than labelling standard photos that are the main subject of computer vision research. Therefore, we suggest to create annotations by exploiting geo-referenced map products such as the CORINE Land Cover data \cite{sumbul2019bigearthnet} and the MOD500 data \cite{schmitt2019sen12ms, he2018detecting}, as well as governmental data \cite{russwurm2018multi}, which contains information relevant to the task one seeks to achieve.
Second, remote sensing images differ significantly in appearance from the close-range images used in the standard literature on scene segmentation \cite{zhu2017deep}. As mentioned before, remote sensing images are usually not with the same high resolution, and multi-spectral remote sensing images come with more bands than conventional photographs. Furthermore, they usually capture large geographical areas with different kinds of land cover, with occlusions, and with illumination changing over time and space. \textcolor{black}{Taking these characteristics into account, downsampling should be avoided to fully exploit the rich information within the remote sensing data. }
\textcolor{black}{Finally}, the specific application scenarios, which in this study is large-scale or even global HSE mapping, should always be taken into account in the whole framework. This means that a spatial split of training and test data should be well designed \cite{geiss2017effect}, and it is not enough to train a model with high test accuracy on a single experimental test set. Instead, the framework should include further applying the trained model on images acquired over all potential regions of interest, for which reasonable accuracy should also be achieved. This requires a robust model in the face of spectral signature changes resulting from social and cultural differences and changing acquisition conditions. 
Therefore, {an} independent accuracy assessment should be carried out in order to comprehensively assess the mapping results\textcolor{black}{. In this way,} a reliable interpretation and understanding of the performance of the framework will be gained.

This paper will present a framework that takes into account the three problems \textcolor{black}{described above, }by fully exploiting state-of-the-art algorithms and techniques, as well as the freely available global satellite images of the Sentinel-2 mission for large-scale HSE mapping. We propose a framework for {large-scale} HSE mapping from Sentinel-2 imagery using deep learning-based approaches \textcolor{black}{with three major parts: 1) preparation of labels and image data, 2) training a well-generalizing semantic segmentation network to learn to map HSE from Sentinel-2 images (Sen2HSE-Net), and 3) a statistically sound accuracy assessment} of the HSE results. This study is intended to provide answers to the following questions: {How can large-scale HSE mapping benefit from CNNs and remote sensing images of medium resolution, in a situation where potentially noisy ground truth data is only available at a regional scale?} \textcolor{black}{How will the network architecture and experimental setup affect the mapping results?} How good are the resulting HSE maps, compared to the existing state-of-the-art products derived at a similar scale?

The remainder of this paper proceeds as follows: Section~\ref{sec:method} elaborates the proposed HSE mapping approach. Section~\ref{sec:exp} details descriptions about the study area and the experimental setup. Section~\ref{sec:res} evaluates the HSE mapping accuracy and visualizes and compares the produced HSE maps to GUF, the GHS built-up grid, {and other datasets from recent studies such as FROM-GLC10,} for several sample test scenes. The following Section~\ref{sec:dis} provides answers to the questions raised above, based on the interpretation and analysis of the achieved results, and discusses the remaining challenges and the possible solutions for the future work. Finally, Section~\ref{sec:conc} summarizes and concludes the work.

\section{HSE mapping with Sen2HSE-Net}
\label{sec:method}

\textcolor{black}{Considering the spatial resolution of available reference data (\SI{20}{\meter}), the sub-pixel geolocation accuracy of Sentinel-2 data \cite{drusch2012sentinel}, as well as the resolution of existing related products (mostly lower than \SI{20}{\meter}), the specific goal of HSE mapping in this study is to detect whether buildings, roads, or other man-made structures are presented{—that is,  larger than 0\%} in a $20\times20$ cell. Using this definition,} the resulting HSE \textcolor{black}{output from Sentinel-2 imagery} will be a binary layer in the Universal Transverse Mercator (UTM) coordinate system, with a ground sampling distance {(GSD)} of \SI{20}{\meter}. This definition is also consistent with the \SI{30}{\meter} Global Human Built-up and Settlement Extent (HBASE) dataset derived from Landsat, which consists of human settlement, built-up areas, and roads \cite{Wang2017}.

\begin{figure}[!tbh]
	\centering
	\includegraphics[width=0.9\textwidth]{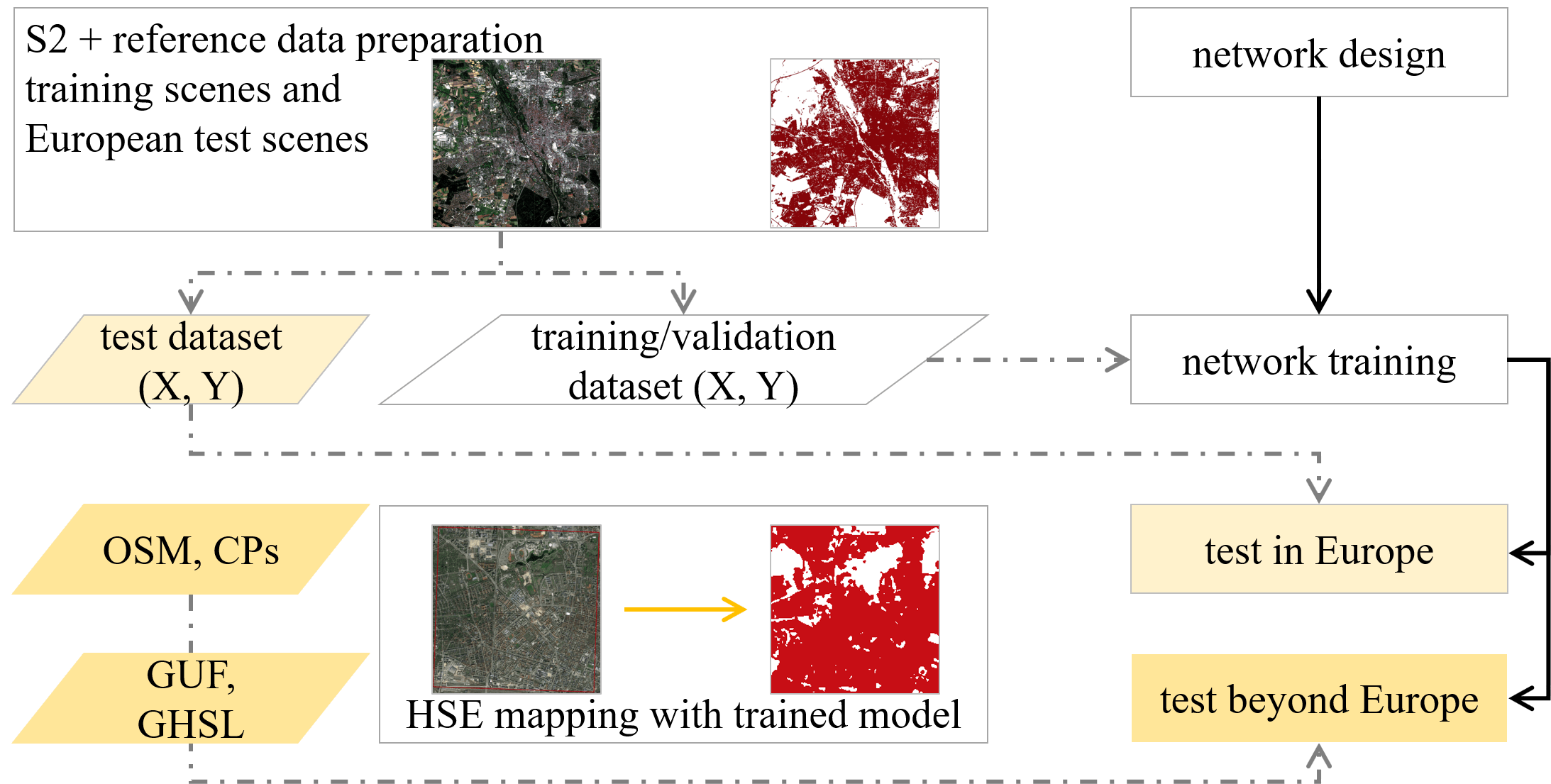}
	\caption{\textcolor{black}{Generalized framework for HSE mapping. The network is instanced as Sen2HSE-Net and compared with several baselines in this study. X and Y are Sentinel-2 image patch and HSE label, respectively.
	}}
	\label{fig:framework}
\end{figure}

The procedure used in the proposed HSE mapping framework is illustrated in Fig. \ref{fig:framework}, which consists of image and reference data preparation, deep neural segmentation network training, and HSE mapping and assessment. Each step will be detailed in the following subsections.

\subsection{Sentinel-2 image pre-processing and reference ground truth preparation}
For each of the cities under study, one (mostly) cloud-free Sentinel-2 image is prepared with Google Earth Engine (GEE) \cite{gorelick2017google}, by exploring a cloud-based engineering approach. The processing approach, described in detail in \cite{Aggregating}, relies on pixel-wise cloud detection and the combination of multi-temporal images \textcolor{black}{within} short time periods. For each study area, we used three Sentinel-2 images compiled from all data acquired for spring, summer, and autumn 2017. The image data contains 13 spectral bands representing Top of Atmosphere Reflectance scaled by a factor of 10000. These images are orthoimages in UTM projection. We used ten of the bands: \textcolor{black}{specifically}, the channels with a GSD of \SI{10}{\meter}, B2 (blue), B3 (green), B4 (red), and B8 (Near-infrared), as well as the \SI{20}{\meter} GSD bands, B5 (red edge 1), B6 (red edge 2), B7 (red edge 3), B8a (red edge 4), B11 (short-wavelength infrared 1), and B12 (short-wavelength infrared 2). In order to create composites with a consistent image size, we up-sampled the second group of bands to a GSD of \SI{10}{\meter} using cubic resampling.
The employed reference data is ``High Resolution Layer Imperviousness 2015,'' an operational product, released as part of the Copernicus Land Monitoring Service's product portfolio \cite{CopernicusHRLI}. ``High Resolution Layer Imperviousness 2015'' is a raster layer indicating built-up areas with a spatial resolution of 20m, created from Copernicus high resolution remote sensing images (mainly the Indian Remote Sensing Satellite and SPOT 5). It is produced using supervised classification, NDVI-based calibration, and subsequent visual improvement. The \textcolor[rgb]{0,0,0}{producer and user} accuracies are supposed to be about 90\%. For registration of reference data and Sentinel-2 images, the reference data is re-projected to the UTM coordinate system and resampled to the extent of the corresponding images.

\begin{figure}[!tbh]
\centering
\subfigure[][Sentinel-2 image]{
\includegraphics[height=0.3\textwidth]{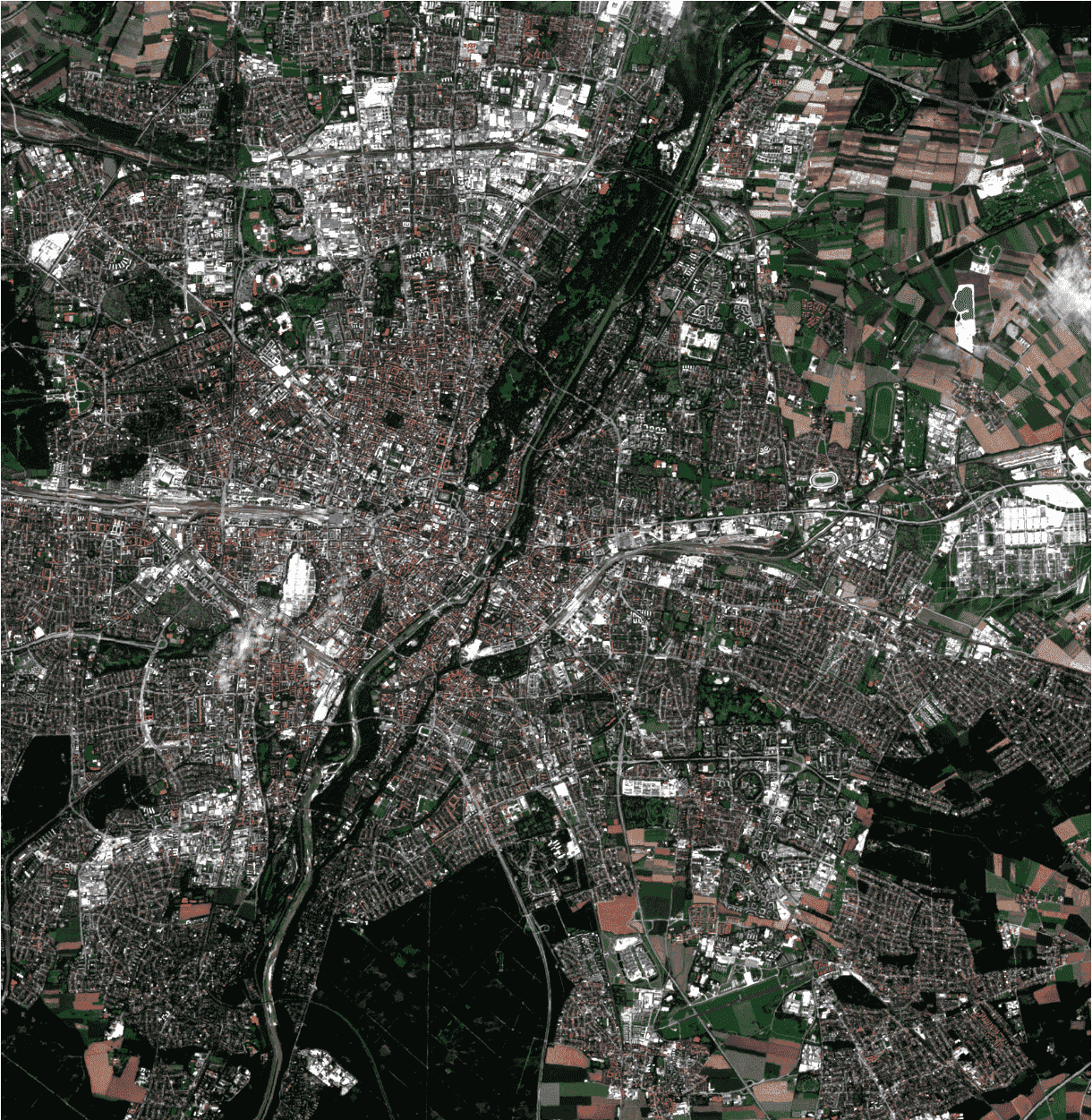}
}
\subfigure[][HSE reference]{
\includegraphics[height=0.3\textwidth]{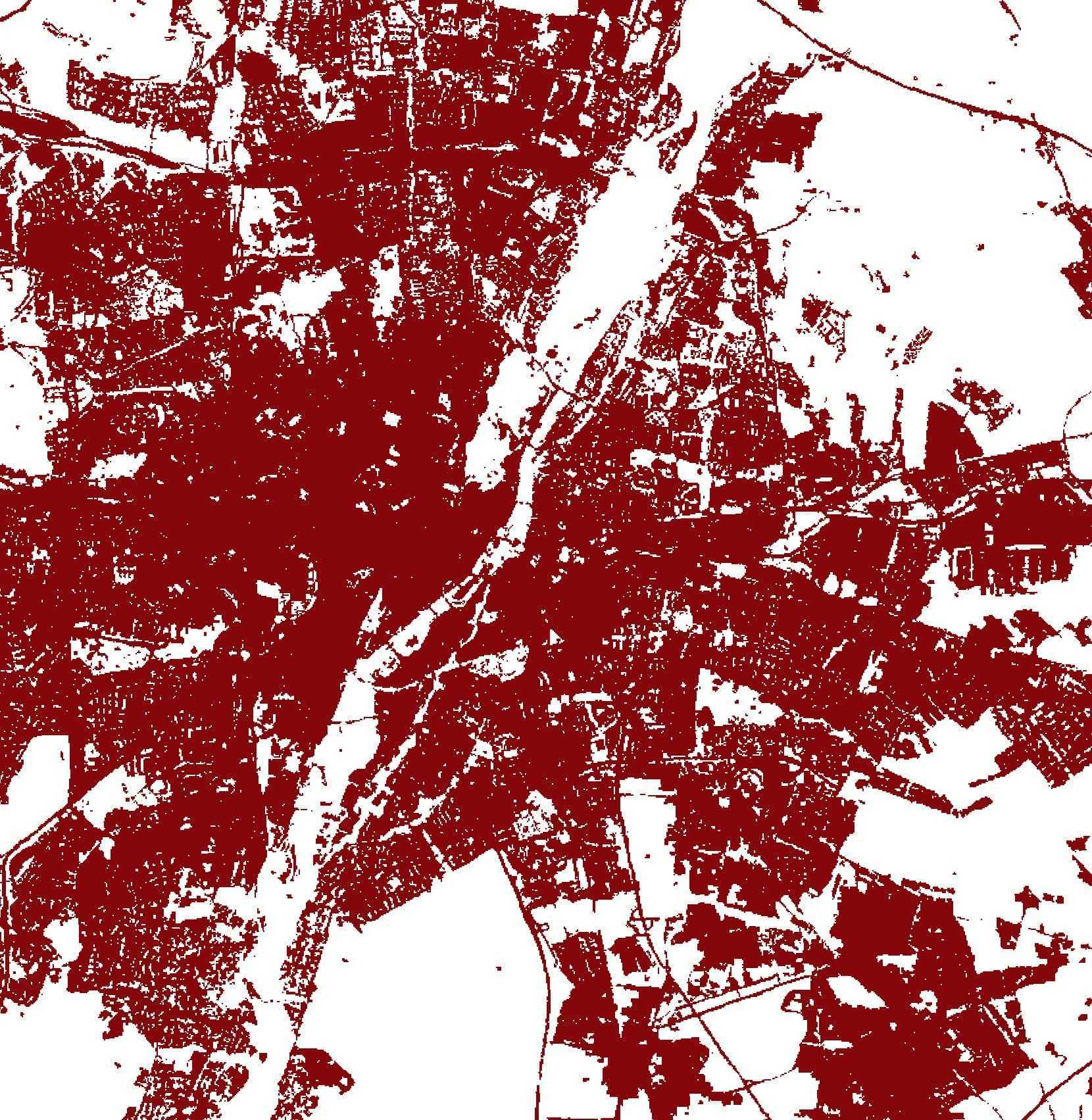}
}
\caption{The processed Sentinel-2 image of central Munich, Germany, and the reference data.
}
\label{fig:imgGt}
\end{figure}

As an example, Fig. \ref{fig:imgGt} illustrates the processed Sentinel-2 image of central Munich, Germany, and the reference data.

\subsection{Convolutional Neural Networks for semantic segmentation}
CNNs currently are the state of the art in visual recognition tasks such as classification and detection, \textcolor{black}{due to} their ability to learn multi-scale representations with high predictive power from example data. They usually consist of basic layers such as convolutional layers composed of weights and biases, pooling layers for a summary of connected activations in feature maps, and activation layers for injecting non-linearity into the models. Some recent examples architectures include forms of the residual convolutional neural network (ResNet), ResNeXt, Inception, and Xception \cite{he2016deep, xie2017aggregated, szegedy2015going, chollet17xception}, among many others. FCNs and their extensions inherit the basic structure of CNNs and replace the fully connected layer\textcolor{black}{, i.e., the last layer in the CNNs, }with a fully convolutional
layer. They feature downsampling (encoder) together with subsequent upsampling (decoder) to maintain the resolution of the input image in the output map.

{There are two approaches for remote sensing image classification via deep learning: \textcolor{black}{working} with \textcolor{black}{either} patch-based CNNs designed for image classification \cite{paisitkriangkrai2016semantic, langkvist2016classification, russwurm2018multi, zhang2019joint, zhong2019deep, hua2019relation, hua2019recurrently, zhu2019so2sat} or encoder-decoder-like neural networks designed for semantic segmentation \cite{maggiori2016fully, maggiori2016convolutional, fu2017classification, volpi2016dense}. \textcolor{black}{The former} works under the assumption of just a single label for each image patch, and applies the trained model to the image of a study area via a sliding window approach\textcolor{black}{,} with the target GSD as the stride of the sliding window. In contrast, \textcolor{black}{the latter approach}, FCNs are designed to predict pixel-level labels, and after training, they can accept inputs of arbitrary size. Their advantages are a potentially higher accuracy resulting from the inter-patch context information (only the intra-patch context is considered in patch-based CNN approaches), and less expensive computation\textcolor{black}{, since} overlapping patches are avoided when using the sliding window method for dense prediction.}

\textcolor{black}{Given both the goal of our task---to assign a label, HSE or non-HSE, to each $20\times20$ meter patch---and the advantages of pixel-level recognition, we decided to combine the patch-based CNN approach and pixel-level recognition approach. Instead of inputting a $20\times20$ meter patch into the network and outputting one label for the patch, we feed larger patches to the network and predict labels for each $2\times2$ pixels by including one pooling (downsampling) layer in the network.}


\subsection{Architecture and training of Sen2HSE-Net}

\begin{figure}[!tbh]
	\centering
	\includegraphics[width=0.99\textwidth]{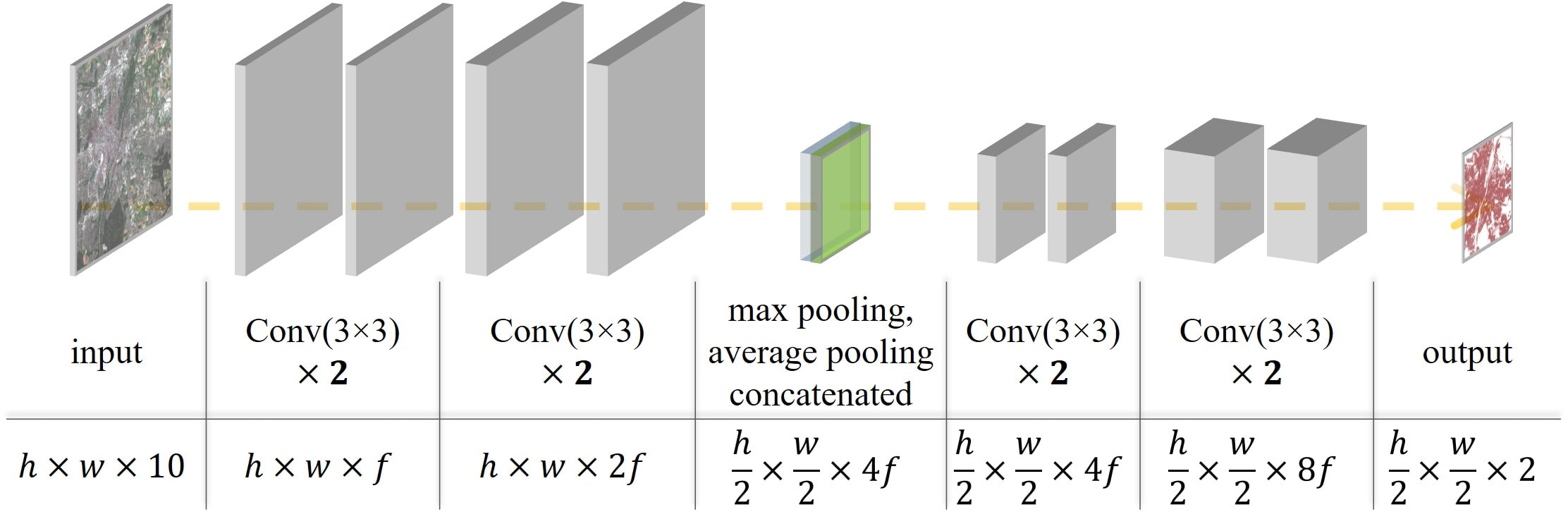}
	\caption{\textcolor{black}{Architecture and details of Sen2HSE-Net. The terms ``h'', ``w'', and ``f'' denote height, width, and the channel number of the first feature maps, respectively. The different size of the final prediction from the input image is due to the different resolution of the HSE prediction {(with a 20-meter GSD)} to the input image {(with a 10-meter GSD)}.
	}}
	\label{fig:network}
\end{figure}

\textcolor{black}{Considering that the network should be kept as simple as possible to make it feasible for reproduction and upscaling, we implemented a simple FCN, the architecture of which is illustrated in Fig. \ref{fig:network}. It consists of four convolutional layers in the beginning to extract low-level features from the input Sentinel-2 images, two pooling layers (maximum and average pooling) in the middle to abstract the learned features to a higher level, then four convolutional layers to extract high-level features, and one convolutional layer in the end for predictions. The kernel sizes for the two sets of four convolutional layers are $3\times 3$; the last convolutional layer has a kernel size of $1\times 1$. Additionally, there are two drop-out layers to avoid model overfitting to the training data, given that the goal is to map HSE globally. No additional pooling layers are used to avoid the information loss during downsampling process, which is also the design idea in \cite{lang2019country} and \cite{hasanpour2016lets}. As defined, the output prediction is with a 20-meter GSD, {while the input data is with a 10-meter GSD}; thus no upsampling layers are used.}

\textcolor{black}{Filter weights are initialized using {the algorithm} 
proposed by \citep{he2015delving}. {The number of output filters of the first convolutional layer, $f$, is set as 16} 
in the experiments and adjusted for investigations in Sec. \ref{sec:dis}. The input images and their corresponding reference labels are used to train the network with the Nesterov Adam optimizer implementation of Keras \cite{chollet2015keras}. We used a minibatch size of 8 images and fixed learning rate of $2\times10^{-4}$. To control the training time and avoid overfitting, early stopping was used, and the monitored metric is the validation loss with patience of 10 epochs, which means that the training stops if the validation loss does not decrease for 10 epochs. All the experiments were carried out using the same setups described above, in order to make for meaningful comparisons.}

\section{Experimental setup}
\label{sec:exp}
\subsection{Study area and training data preparation}
\begin{figure}[!tbh]
	\centering
	\includegraphics[width=0.99\textwidth]{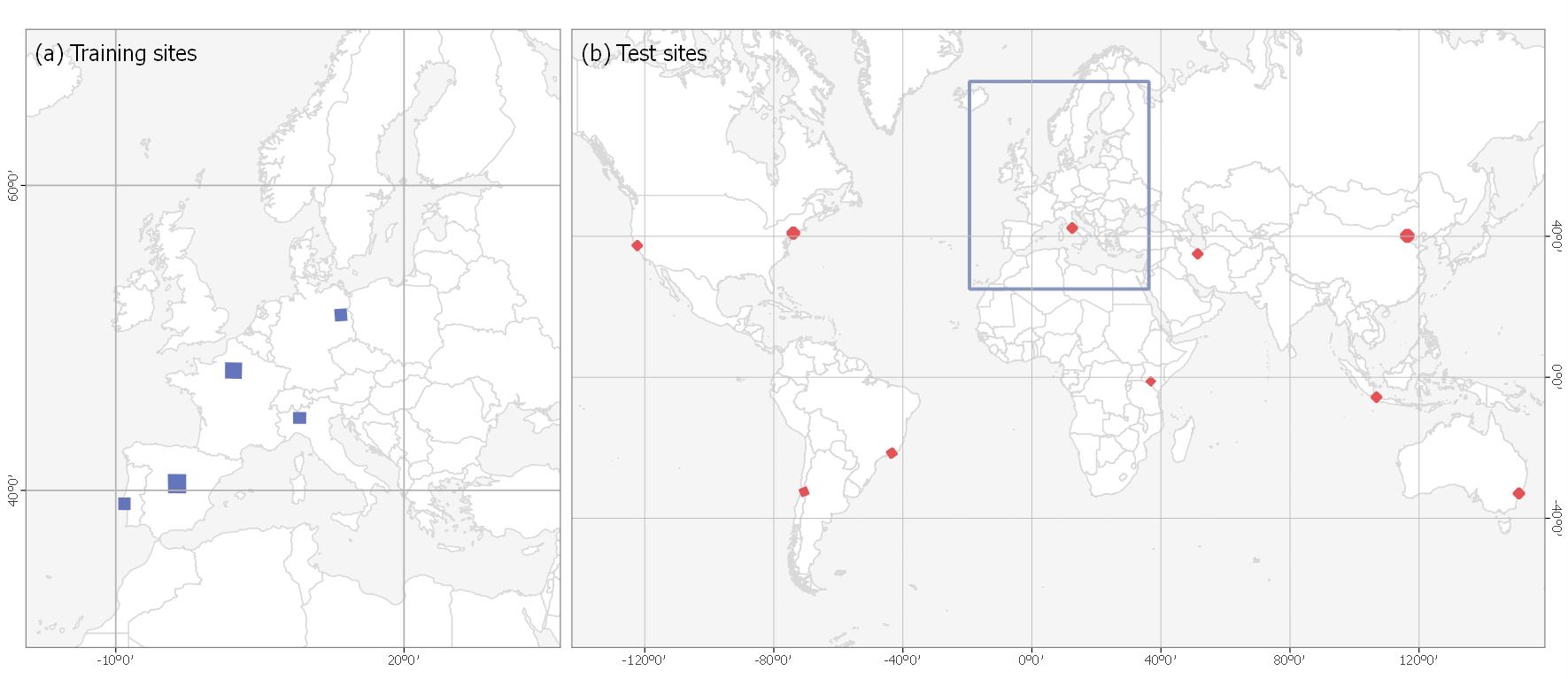}
	\caption{Five training areas distributed across Europe and {ten} test areas  across the world.}
	\label{fig:7city}
\end{figure}

The training areas are five cities in Central Europe, as shown in Fig. \ref{fig:7city}. These cities are chosen for training because the reference ground truth data is only available in Europe. The test areas are ten cities across the world, as shown in Fig.~\ref{fig:7city}. \textcolor{black}{In addition to these ten test scenes distributed across the world, three test scenes in Europe are also chosen to provide a basis for evaluating the regional-to-global generalization capability of the proposed framework.} Table \ref{tab:cityChrac} describes the main characteristics of the selected test cities, which differ in urban area, topography, and land-cover features in the surrounding countryside.

\begin{table}[!tbh]
\scriptsize
  \centering
  \caption{Basic information of the study areas for training and test, and the urban ecoregions according to \cite{Schneider.2010}. }
    \begin{tabular}{llll}
    \toprule
    &city & urban ecoregion & { area ($km^2$)}\\
    \midrule
         \multirow{5}[1]{*}{training scenes }&
 Berlin, Germany & Temperate forest in Europe & 5138
 \\   
   & Lisbon, Portugal & Temperate mediterranean & 4585
 \\  
  &  Madrid, Spain & Temperate mediterranean & 19360
 \\ 
  &  Milan, Italy & Temperate mediterranean & 5512
 \\  
  &  Paris, France & Temperate forest in Europe & 11561
 \\   
  \midrule
   \multirow{3}[1]{*}{European test scenes}&
    Amsterdam, Netherlands & Temperate forest in Europe & 9714
 \\  
  & London, England & Temperate forest in Europe & 6711
  \\  
  & Munich, Germany & Temperate forest in Europe & 7355
 \\     
    \midrule
    \multirow{10}[1]{*}{test scenes beyond Europe}&
    Beijing, China & Temperate forest in East Asia & 11017 \\
   & Nairobi, Kenya & Tropical, sub-tropical savannah in Africa & 591 \\
   & Rome, Italy & Temperate mediterranean & 2890 \\
   & {Rio de Janeiro, Brazil} & Tropical, Sub-tropical savannah in South America  & 2492 \\ 
   & San Francisco (SF), USA & Temperate mediterranean & 1784 \\
   & Santiago, Chile & Temperate mediterranean & 2890 \\
   & Sydney, Australia & Temperate forest in North America & 1894 \\
   & Tehran, Iran & Temperate grassland in Middle East Asia & 1678 \\
   & {Jakarta, Indonesia} &   Tropical, Sub-tropical forest in Asia & 2492 	\\ 
   & {New York City (NYC), USA} &  Temperate forest in North America & 7355 \\ 
    \bottomrule
    \end{tabular}%
  \label{tab:cityChrac}%
\end{table}%

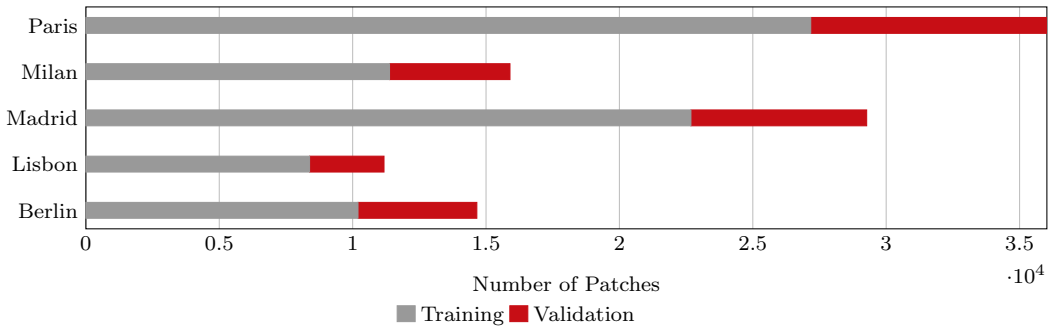
\begin{figure}
	\centering

	\begin{tikzpicture}
		\scriptsize

    	\pgfplotstableread{
            label 		X1   num1  num2
            {Berlin}	1	10200	4455
            {Lisbon}	8	8385	2790
            {Madrid}	15	22680	6588
            {Milan}	    22	11382	4515
            {Paris}	    29	27171	8853
            }\data

    	\begin{axis}[
    	    width=.95\linewidth,
    	    height=0.3\linewidth,
        	xbar stacked,
        	bar width=6pt,
        	yticklabels from table={\data}{label},
            ytick=data,
            ytick style={draw=none},
            xlabel={Number of Patches},
            xmin=0,
            xmax=36024,
        	xtick style={draw=none},
            xmajorgrids,
            legend style={cells={anchor=west},
        			at={(0.45,-0.3)},
        			anchor= north,
        			legend columns=2,
        			draw=none},
    	    ]
        	\addplot+ [color=tumgray] table [x=num1, y expr=\coordindex] \data;
        	\addplot+ [color=compMR] table [x=num2, y expr=\coordindex] \data;
        	\legend{Training, Validation}
    	\end{axis}
	\end{tikzpicture}

	\caption{\textcolor{black}{Number of training and validation patches in our dataset.}}
    \label{fig:trainPatch}
\end{figure}

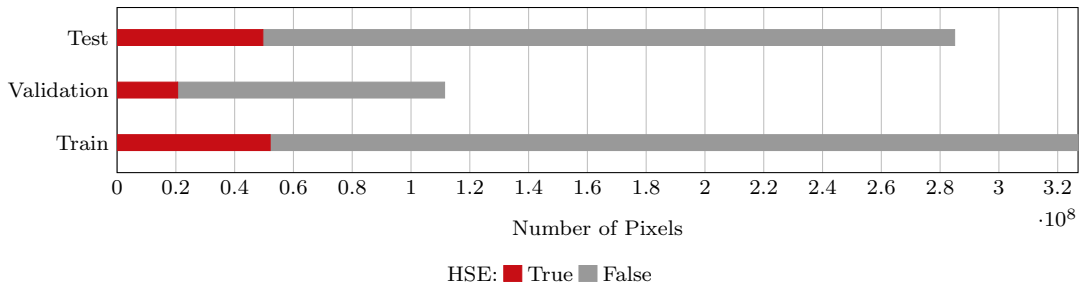
\begin{figure}
	\centering

	\begin{tikzpicture}
		\scriptsize

    	\pgfplotstableread{
            label X1   num1  num2
            {Train}	40 52069215 274865313
            {Validation} 110 20595333 90819963
            {Test}	180 49581801 235356439
            }\data

    	\begin{axis}[
    	    width=.95\linewidth,
    	    height=0.25\linewidth,
        	xbar stacked,
        	bar width=6pt,
        	yticklabels from table={\data}{label},
            ytick=data,
            ytick style={draw=none},
            xlabel={Number of Pixels},
            xmin=0,
            xmax=326934528,
            ymin=0,
            ymax=220,
        	xtick style={draw=none},
            xmajorgrids,
            legend style={cells={anchor=west},
        			at={(0.45,-0.5)},
        			anchor= north,
        			legend columns=3,
        			draw=none},
    	    ]
             \addlegendimage{empty legend}
             \addplot+ [color=compMR] table [x=num1, y=X1] \data;
             \addplot+ [color=tumgray] table [x=num2, y=X1] \data;
        	 \legend{{HSE}:, True, False}
    	\end{axis}
	\end{tikzpicture}
	\caption{\textcolor{black}{Number of pixels in training, validation, and test datasets. The test data presented here is from the three scenes in Europe.}}
    \label{fig:testPixel}
\end{figure}

After coregistration, HSE reference data and Sentinel-2 images were cropped into patches of \SI[product-units = single]{128x128}{\pixel} with a stride of \SI{96}{\pixel}. 
The final patches were spatially split into a training and a validation subset. \textcolor{black}{The exact number of patches from each training scene is presented in Fig. \ref{fig:trainPatch}. The number of HSE and non-HSE pixels in the training, validation, and test datasets in Europe is presented in Fig. \ref{fig:testPixel}.}

\subsection{Accuracy assessment strategy}

\begin{figure}[htbp]
	\centering
	\includegraphics[width=0.85\textwidth]{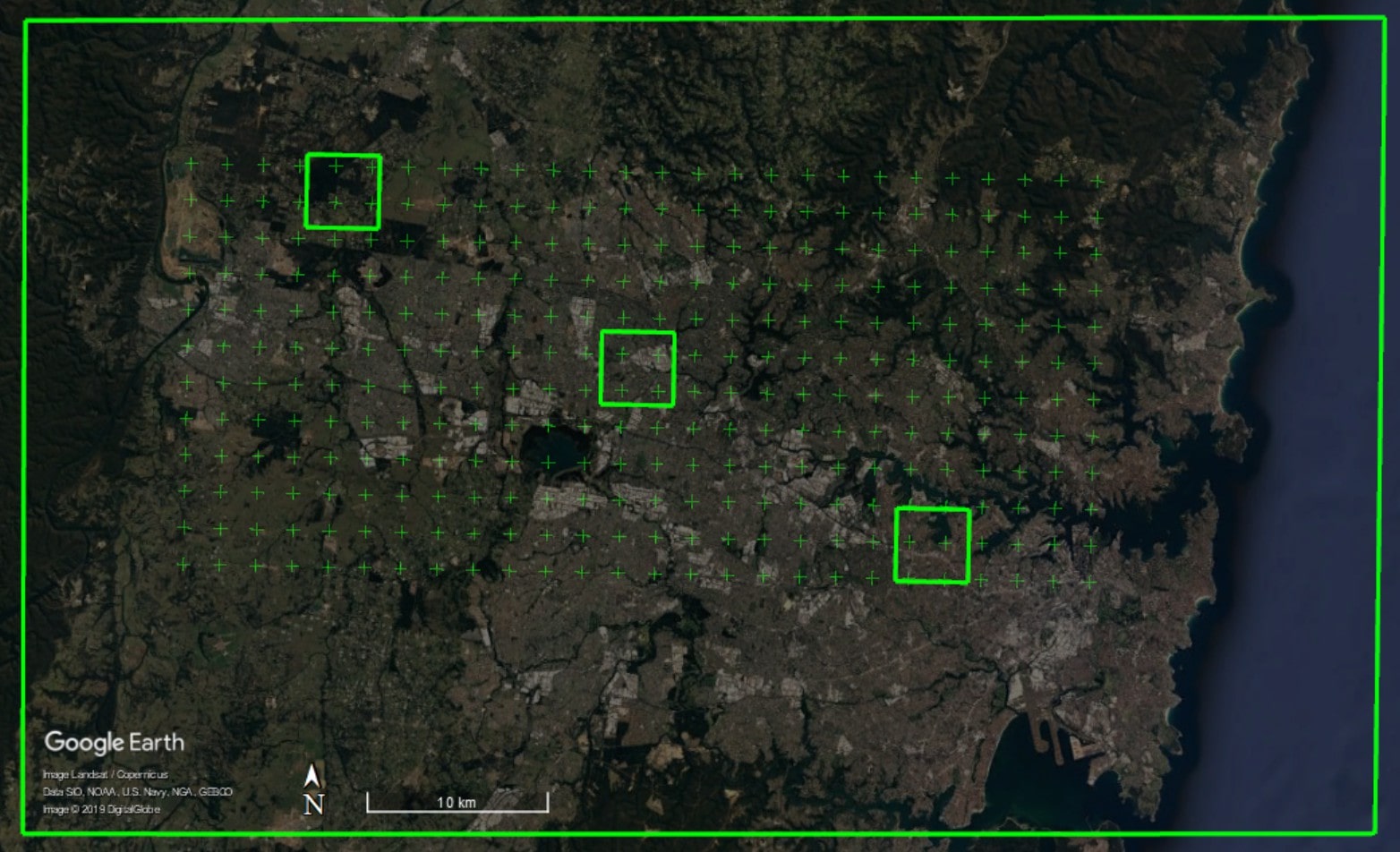}
	\caption{The MLGCPs and three subset regions for closer visualization within the ROI, using the city of Sydney as an example. A similar configuration is used for the assessment of all test cases.}
	\label{fig:checkGrid}
\end{figure}

\begin{figure}
	\centering
    \begin{tikzpicture}
    	\scriptsize

    	\pgfplotstableread{
            label 		X1   num1  num2
            {Beijing}	1	172	185
            {Nairobi}	8	78	87
            {Rome}	15	93	162
            {Rio}	22	51	153
            {Sanfrancisco}	29	142	112
            {Santiago}	36	166	146
            {Sydney}	43	113	96
            {Tehran}	50	312	168
            {Jakarta}	57	182	418
            {New York}	64	363	209
            }\data

    	\begin{axis}[
    	    width=.95\linewidth,
    	    height=0.4\linewidth,
        	xbar stacked,
        	bar width=6pt,
        	yticklabels from table={\data}{label},
            ytick=data,
            ytick style={draw=none},
            xlabel={Number of Samples},
            xmin=0,
            xmax=600,
        	xtick={0, 200, ..., 600},
        	xtick style={draw=none},
            xmajorgrids,
            legend style={cells={anchor=west},
        			at={(0.45,-0.3)},
        			anchor= north,
        			legend columns=3,
        			draw=none},
    	    ]
    	    \addlegendimage{empty legend}
        	\addplot+ [color=compMR] table [x=num1, y expr=\coordindex] \data;
        	\addplot+ [color=tumgray4] table [x=num2, y expr=\coordindex] \data;
        	\legend{{HSE}:, True, False}
    	\end{axis}
    \end{tikzpicture}
    \caption{\textcolor{black}{Number of MLGCPs for HSE mapping assessment. A different number of points are chosen for different cities to ensure diversity in land covers by including different city areas.}}
    \label{fig:dataTestHSE}
\end{figure}
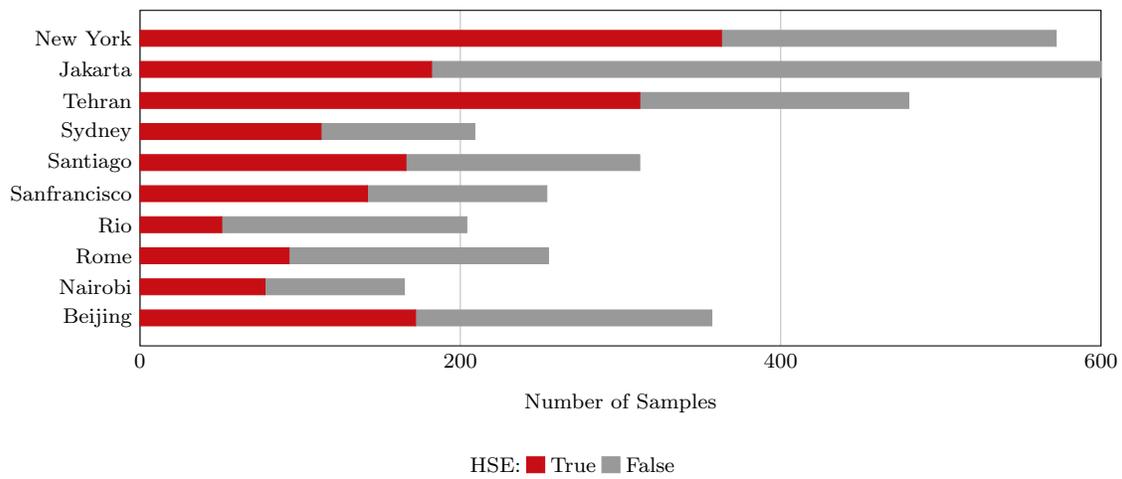

Manually labeled ground truth is employed for {a} \textcolor[rgb]{0,0,0}{quantitative assessment}. In order to avoid human-induced bias, an equally distributed grid is generated for each test city, in the city center area, with 2000 meters distance between each point. These manually labeled grid-based checking points (MLGCPs), with a size of $20m\times 20m$, are manually classified into HSE or non-HSE. This fixed distribution of check points allows for a meaningful spatial assessment of the mapping results. For similar reasons, three fixed subset regions, with a size of $4km\times4km$, distributed across the whole region of interest (ROI), are chosen for each city for a closer view of the produced results. Figure \ref{fig:checkGrid} illustrates the three subset regions and the MLGCPs within the ROI, using Sydney as an example. The number of test samples of all ten test scenes is presented in Fig. \ref{fig:dataTestHSE}.

Furthermore, several state-of-the-art products were chosen for comparison based on the following riteria: they should be available on a global scale, be provided with a similar pixel spacing, and provide relevant information about HSE, because only similar characteristics enable an extensive and consistent comparison. Therefore, we chose GUF, the GHS built-up grid, FROMGLC10
, and High-resolution Multi-temporal Global Urban Land (HMGUL) \cite{liu2018high} as the baselines for comparison and validation of the HSE mapping results produced by our approach. The details of these reference products are provided in Tab. \ref{tab:refData}. All baseline products were re-sampled to 20 m GSD for comparison with the produced maps in this study. For the purpose of comparison, the ``built-up'' and ``built-up up to 2014'' are taken from GUF and GHSL as the HSE information, respectively. Because \textcolor{black}{neither} of these products should be considered as ground truth, as they were all created by different mapping approaches, we do not test our results against them. Instead, we compare our results to these datasets with respect to independent references.

\begin{table}
    \centering
    \footnotesize
    
    \caption{Description of baseline products for accuracy comparison.}
    \begin{tabularx}{\linewidth}{@{}XlXlX@{}}
        \toprule
        Product & Sensor & Year & Label & GSD \\
        \cmidrule(r){1-1} \cmidrule(lr){2-2} \cmidrule(lr){3-3} \cmidrule(lr){4-4} \cmidrule(l){5-5}
        GUF & TerraSAR/TanDEM-X & \numrange{2011}{2014} & built-up, with vertical component & \SI{12}{\meter} \\
        GHSL & Landsat & \numrange{1975}{2014} & multi-epoch built-up grid & \SI{38}{\meter} \\
         FROMGLC & Sentinel-2 & {2017} & impervious surface & \SI{10}{\meter} \\
         HMGUL & Landsat & {2015} & urban & \SI{30}{\meter} \\
    \bottomrule
    \end{tabularx}
    
  \label{tab:refData}
\end{table}

In addition to {quantitative and visual comparisons} with similar products, {a} quantitative assessment is also performed with respect to the OpenStreenMap building layer used as the ground truth reference. Because the mapped HSE includes not only buildings but also other man-made structures, such as roads, we only employed recall as the indicator\textcolor{black}{. That is} $recall= \dfrac{N_1}{N_0}$, where $N_0$ is the number of all building pixels based on OSM, and $N_1$ is the number of pixels (in $N_0$) also mapped as HSE. This way, we are aiming at the detection rate of buildings in the mapping results. A good HSE map should include all buildings provided in the OSM building layer. It should be mentioned that the quality of the crowdsourced OSM reference data is not homogeneous over the cities and the suburban areas, as well as over developing and developed countries, in terms of completeness and thematic accuracy  \cite{fan2014quality, arsanjani2015quality, johnson2017employing, viana2019value}. Therefore, in our study, the OSM-based evaluation results are only provided as an additional rough accuracy estimate of the HSE mapping results and should be primarily used \textcolor{black}{for a relative comparison of the results}. Additionally, buildings are also included in both the GHS built-up grid and GUF datasets, according to their definitions. Therefore, the detection power of these two layers is also presented by the above defined recall metric for comparison\textcolor{black}{, in order} to gain an intuitive estimation of the quality of the mapped HSE.

\section{HSE mapping results}
\label{sec:res}
The results of the experimental assessment of the proposed HSE mapping framework are illustrated in this section. First, accuracy assessments with respect to different reference data are shown. We \textcolor{black}{then} compile the comparison between the mapped HSE and the \textcolor{black}{state-of-the-art} products for several cities across the world. For better evaluation, we visualize the comparison at both the city scale and building block scale. \textcolor{black}{Finally}, case studies for large-scale HSE mapping are provided to demonstrate the upscaling potential of the proposed framework.

\subsection{Quantitative assessment of HSE mapping results}
\label{sec:accManully}

For the ten globally distributed cities, accuracy assessments are carried out with two kinds of reference data, MLGCPs and OSM. The kappa coefficient, average accuracy (AA) of the two classes (HSE and non-HSE),  commission error, recall\textcolor{black}{,} and \textcolor{black}{F-Score of HSE} are shown in Tab. \ref{tab:MLGCPs_osm_cmp}.  To \textcolor{black}{provide} a sense of the quality of the achieved results, we also list the corresponding assessment results for the state-of-the-art products, GUF and GHS.

\begin{table}[!tbh]
\scriptsize
  \centering
  \caption{\textcolor{black}{Accuracy assessment of HSE mapping results from Sen2HSE-Net by kappa, AA (in percentage), commission error (CME, in percentage), recall (in percentage), and F-Score with respect to the MLGCPs and OSM reference data. The corresponding assessment of GUF and the GHS built-up grid is also listed for comparison. Only recall with respect to OSM is presented, given the different definitions of HSE and OSM reference.}}
    \begin{tabular}{cclccccccccccc}
    \toprule
    \multicolumn{2}{c}{reference} & source & Beijing & Nairobi & Rome  & Rio   & SF    & Santiago & Sydney & Tehran & Jakarta  & NYC & \textbf{Mean} \\
    \multirow{12}[7]{*}{MLGCPs} & \multirow{3}[1]{*}{Kappa} & ours  & \textbf{0.75} & \textbf{0.73} & \textbf{0.79} & \textbf{0.88} & \textbf{0.88} & \textbf{0.90} & 0.78  & 0.67  & \textbf{0.89} & \textbf{0.82} & \textbf{0.81} \\
          &       & GUF   & 0.64  & 0.70  & 0.75  & 0.81  & 0.81  & 0.75  & \textbf{0.81} & \textbf{0.76} & 0.60  & 0.62  & 0.73 \\
          &       & GHSL   & 0.54  & 0.37  & 0.77  & 0.74  & 0.87  & 0.65  & 0.77  & 0.70  & 0.36  & 0.72  & 0.65 \\
\cmidrule{2-14}          & \multirow{3}[2]{*}{AA} & ours  & \textbf{87.6} & \textbf{86.1} & \textbf{88.2} & \textbf{94.4} & 93.8  & \textbf{94.8} & 88.6  & 81.8  & \textbf{94.4} & \textbf{91.0} & \textbf{90.1} \\
          &       & GUF   & 81.9  & 84.7  & 85.9  & 89.0  & 88.9  & 88.4  & \textbf{91.0} & \textbf{88.5} & 80.7  & 83.9  & 86.3 \\
          &       & GHSL   & 77.2  & 68.0  & 87.8  & 90.1  & \textbf{95.4} & 83.7  & 88.4  & 85.6  & 65.9  & 85.3  & 82.7 \\
\cmidrule{2-14}          & \multirow{3}[2]{*}{CME} & ours  & 17.4  & 4.8   & 6.3   & 9.6   & \textbf{6.8} & 5.9   & 15.8  & 15.1  & \textbf{8.2} & 6.6   & 9.7 \\
          &       & GUF   & \textbf{16.3} & 3.4   & \textbf{5.5} & \textbf{5.8} & 8.9   & \textbf{3.4} & \textbf{4.6} & \textbf{6.0} & 12.7  & \textbf{4.3} & \textbf{7.1} \\
          &       & GHSL   & 26.1  & \textbf{3.3} & 11.6  & 26.9  & 14.0  & 4.7   & 13.8  & 9.0   & 26.3  & 11.7  & 14.7 \\
\cmidrule{2-14}          & \multirow{3}[2]{*}{recall} & ours  & \textbf{93.6} & \textbf{75.6} & 79.6  & 92.2  & \textbf{96.5} & \textbf{96.4} & \textbf{99.1} & \textbf{95.2} & 92.3  & \textbf{93.4} & \textbf{91.4} \\
          &       & GUF   & 77.9  & 71.8  & 74.2  & 80.2  & 80.4  & 80.3  & 86.7  & 83.2  & 84.0  & 73.6  & 79.2 \\
          &       & GHSL   & 80.8  & 37.2  & \textbf{81.7} & \textbf{95.6} & 96.1  & 71.8  & 94.0  & 80.5  & \textbf{94.2} & 91.7  & 82.4 \\
\cmidrule{2-14}          & \multirow{3}[2]{*}{F-Score} & ours  & \textbf{0.88} & \textbf{0.84} & \textbf{0.86} & \textbf{0.91} & \textbf{0.95} & \textbf{0.95} & \textbf{0.91} & \textbf{0.90} & \textbf{0.92} & \textbf{0.93} & \textbf{0.91} \\
          &       & GUF   & 0.81  & 0.82  & 0.83  & 0.87  & 0.85  & 0.88  & \textbf{0.91} & 0.88  & 0.86  & 0.83  & 0.85 \\
          &       & GHSL   & 0.77  & 0.54  & 0.85  & 0.83  & 0.91  & 0.82  & 0.90  & 0.85  & 0.83  & 0.90  & 0.82 \\
    \midrule
    \multirow{3}[2]{*}{OSM} & \multirow{3}[2]{*}{recall} & ours  & \textbf{97.9} & \textbf{92.2} & \textbf{93.8} & 93.3  & \textbf{99.1} & \textbf{99.1} & \textbf{97.8} & \textbf{97.6} & \textbf{98.1} & 97.7  & \textbf{96.7} \\
          &       & GUF   & 89.7  & 84.2  & 90.1  & 84.1  & 77.6  & 91.2  & 87.0  & 87.8  & 81.5  & 90.1  & 86.3  \\
          &       & GHSL  & 92.9  & 72.6  & 92.1  & \textbf{97.3} & 98.0  & 72.2  & 96.9  & 73.0  & 96.4  & \textbf{97.9} & 88.9 \\
    \bottomrule
    \end{tabular}%
  \label{tab:MLGCPs_osm_cmp}
\end{table}

\textcolor{black}{Table \ref{tab:MLGCPs_osm_cmp} indicates that the achieved HSE mapping results are promising, as they provide the highest kappa, AA, recall, and F-Score on average over ten test scenes, when compared to both of the baseline products. In particular, we achieve the highest F-Score (with respect to the MLGCPs) for all ten distinct test areas across the world. In addition, more buildings (from the OSM layer) are included in the mapping results, compared to both GUF and the GHS built-up grid. This can be seen from the improved mean recall, from 86.3\% and 88.9\% to 96.7\%, compared to GUF and the GHS built-up grid, respectively. This improvement is apparent for eight of the ten cities.}

\textcolor{black}{The commission error from our mapping results, however, is relatively high, especially when compared to GUF, which means that the HSE is overestimated in our results. On the one hand, this shows that GUF is strong at excluding non-HSE from HSE. On the other hand, it is also due to the different mapping focus (vertical artificial structures) of GUF. Still, even considering commission error, our results are generally better than the GHS built-up grid, which is closer to our mapping focus. The GHS built-up grid provides the highest recall in three test scenes with respect to the MLGCPs and two test scenes with respect to OSM. The differences among these three results will be further analyzed in the discussion section. Considering the varying characteristics of the three layers, it should be mentioned that the comparison presented in Tab. \ref{tab:MLGCPs_osm_cmp} is not intended to rank their quality, but rather to provide a validation reference for our mapping results through comparisons.}

\textcolor{black}{The presence of fewer outliers in the representative test scenes shows the good generalization ability and the robustness of the trained model. However, the achieved results do reveal differences among different test scenes. For instance, the result in Nairobi is worse than the average for all three dataset. This is probably due to different urban structures and surrounding terrains, and is indicative of  the challenges for large-scale mapping.}

\subsection{Qualitative assessment of HSE mapping results}
\begin{figure}[!tbh]
\centering
\includegraphics[height=0.9\textheight]{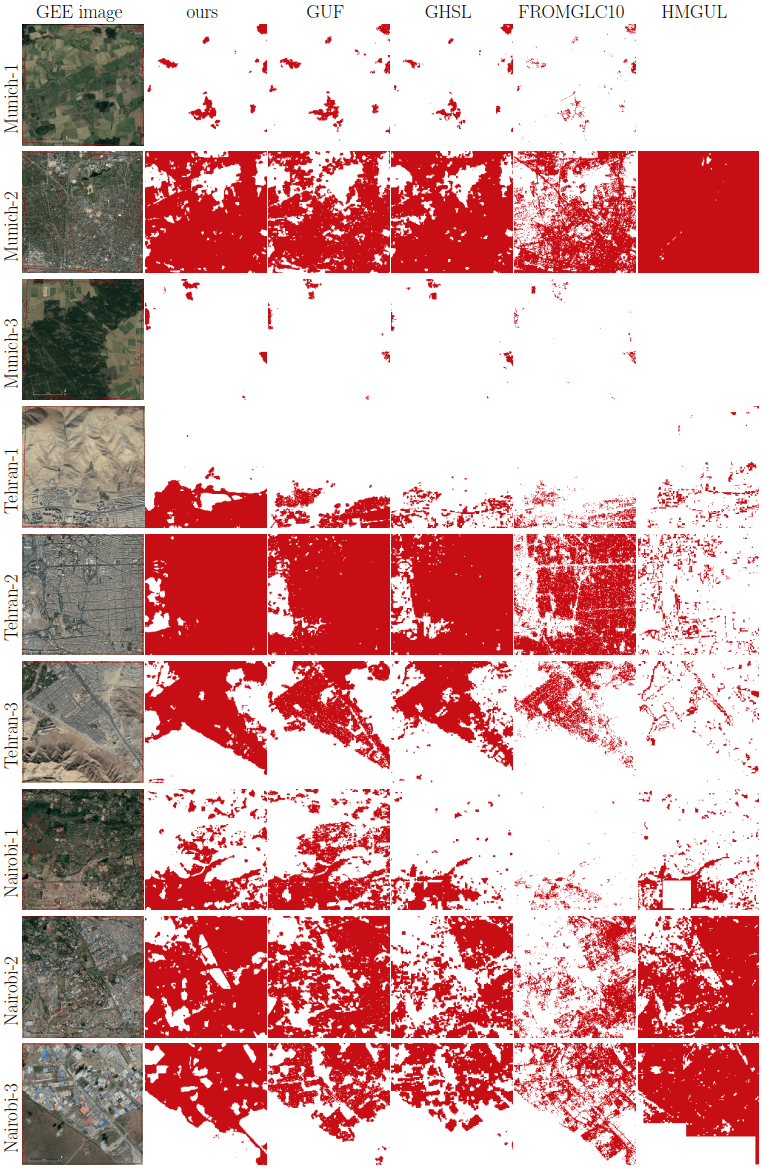}
\caption{{Comparison of the produced HSE results to four state-of-the-art products for three pre-defined subsets in three test scenes.}}
\label{fig:cmp_2}
\end{figure}

The comparison of the produced HSE maps to the state-of-the-art products can be found in Fig. \ref{fig:cmp_2} for the three subset test areas in Munich, Nairobi, and Tehran. Overall, the mapped HSE results are in agreement with the GHS built-up grid, GUF, {and FROM-GLC10, while the HMGUL is in a relative coarse resolution}. \textcolor{black}{From the comparison, it can also be seen that the mapped HSE does include roads, streets, in addition to buildings, as expected. Some roads are also included in the GHS layer, FROM-GLC10, and HMGUL.}

Some superiority of the mapping results can be observed from Fig. \ref{fig:cmp_2}. For instance, the mapped HSE is able to \textcolor{black}{exclude the park area within the city}, as illustrated by the second Munich subset. Also, it is able to include \textcolor{black}{small buildings surrounded by vegetation as well as GUF does}, while the GHS built-up grid {and FROM-GLC10} omit most of the buildings, as illustrated by the first Nairobi subset and the first Munich subset. Additionally, the proposed approach is not affected by the shadow areas of the mountains, which result in false positive results in the GHS built-up grid, as can be seen in the third Tehran subset. 

For a city-scale evaluation of the mapped HSE, \textcolor{black}{the similarities and differences} from the GHS built-up grid and GUF are shown in Fig. \ref{fig:cmp_diff} for \textcolor{black}{three representative test scenes}. \textcolor{black}{The visualization can be interpreted using Tab. \ref{tab:interpretationColor}.} The closer view of the \textcolor{black}{three pre-defined subset regions (as described in Sec. \ref{sec:exp}) of six sample test scenes in Beijing, Nairobi, Rome, San Francisco, Santiago, and Sydney, are shown  in Fig. \ref{fig:cmp_diff_subset}, where high resolution images are also presented for a detailed interpretation}.
 
Figure \ref{fig:cmp_diff} visualizes the overall consistency and agreement of the produced HSE with respect to the GHS built-up grid and GUF. From the test cases in Beijing and Sydney shown in Fig. \ref{fig:cmp_diff}, it can be seen that the main part of a city can be detected by all three datasets, with the urban morphology being shown clearly. The test in Nairobi shows obvious disagreement among these three datasets, which is also noticeable in the other test scenes and will be further analyzed in the discussion section. Figure \ref{fig:cmp_diff} qualitatively shows the general feasibility of the proposed HSE mapping framework and can be further confirmed by the closer view in Fig. \ref{fig:cmp_diff_subset}. By comparing the high resolution images in Fig. \ref{fig:cmp_diff_subset}, we can see that in general our results are able to provide a compact boundary between HSE and non-HSE under a variety of environments in cities across the world. \textcolor{black}{A detailed analysis of this visualization will be presented in the discussion section, providing more evidence of the outstanding performance of the proposed framework.}

\begin{figure*}[!tbh]
\centering
\includegraphics[width=0.8\textwidth]{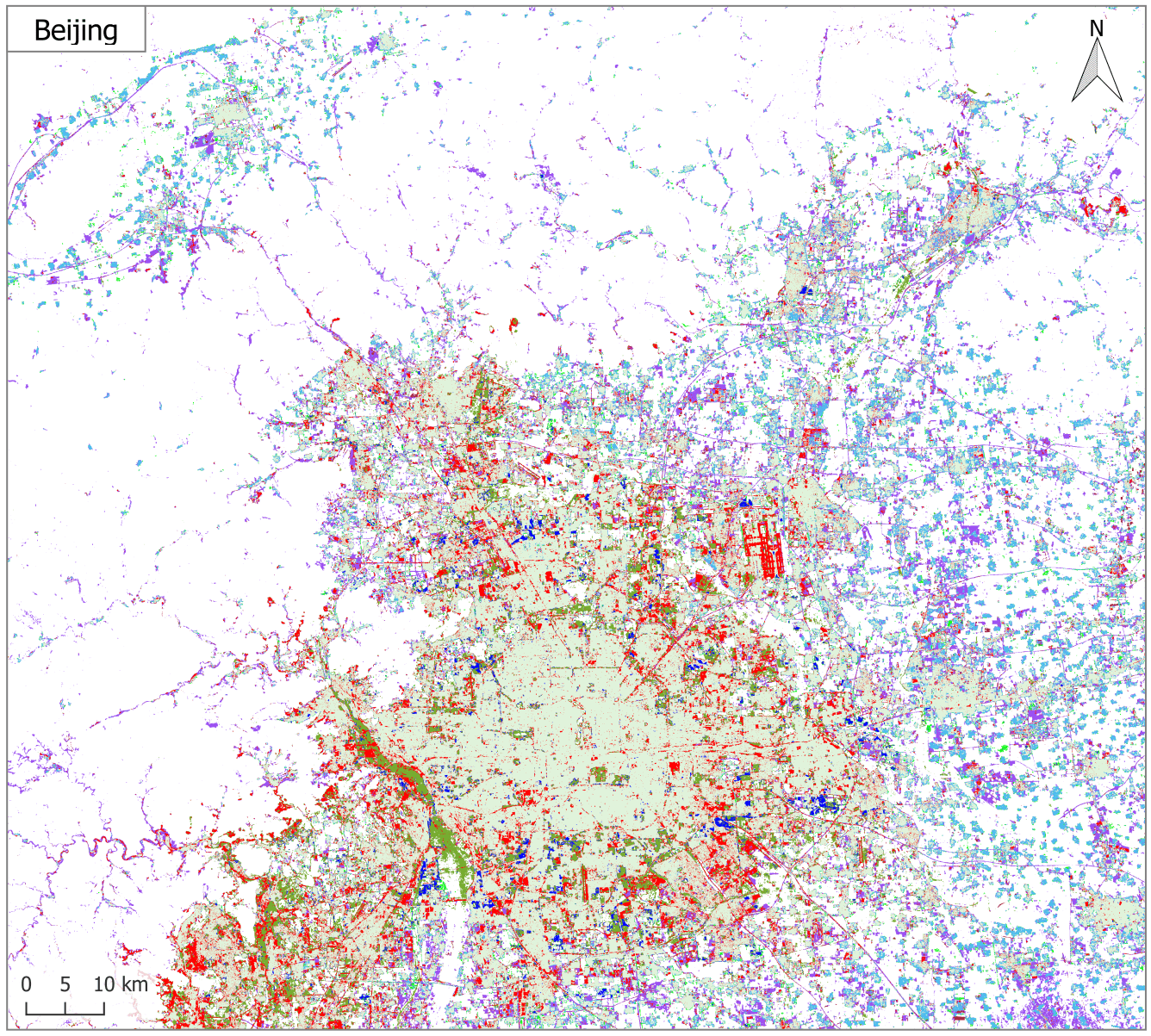}
\begin{tikzpicture}
\pgfplotsset{
legend style={cells={anchor=west}, draw=none,column sep=1ex, nodes={scale=0.6, transform shape}}
}
\begin{customlegend}[legend columns=3]
\addlegendimage{p, only marks, mark=square*}
\addlegendentry{mapped only by our results}
\addlegendimage{gugs, only marks, mark=square*}
\addlegendentry{not mapped only by our results}

\addlegendimage{gu, only marks, mark=square*}
\addlegendentry{mapped only by GUF}
\addlegendimage{gsp, only marks, mark=square*}
\addlegendentry{not mapped only by GUF}

\addlegendimage{gh, only marks, mark=square*}
\addlegendentry{mapped only by GHSL built-up grid}
\addlegendimage{gup, only marks, mark=square*}
\addlegendentry{not mapped only by GHSL built-up grid}

\addlegendimage{b3, only marks, mark=square*}
\addlegendentry{mapped by all 3 datasets}
\end{customlegend}
\end{tikzpicture}
\caption{Produced HSE maps of three representative test scenes, compared to the reference GUF and GHSL built-up grid datasets.}
\label{fig:cmp_diff}
\end{figure*}

\begin{figure*}[!tbh]\ContinuedFloat
\centering
\includegraphics[width=0.8\textwidth]{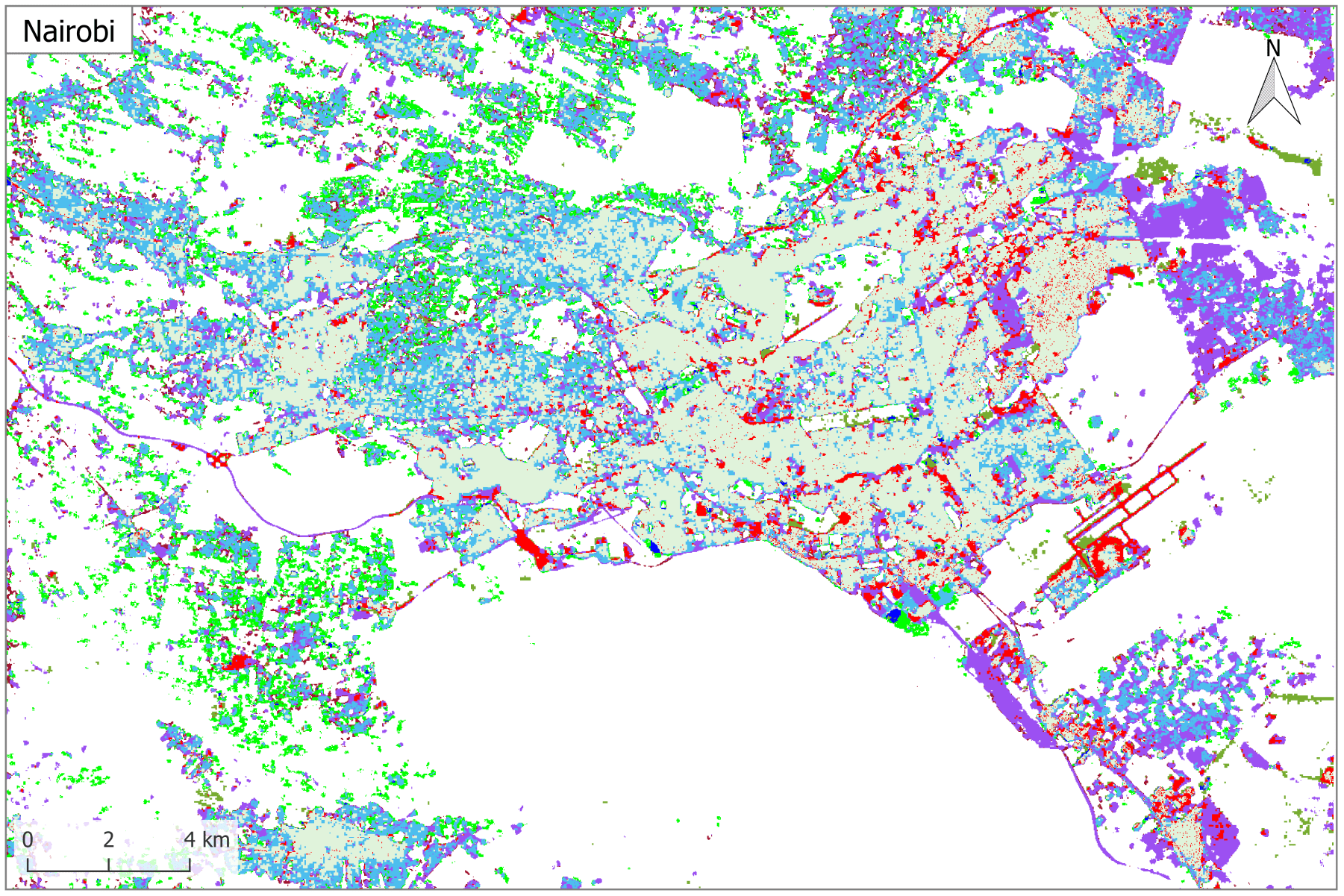}
\caption{Continued.}
\label{fig:cmp_diff}
\end{figure*}
\begin{figure*}[!tbh]\ContinuedFloat
\centering
\includegraphics[width=0.8\textwidth]{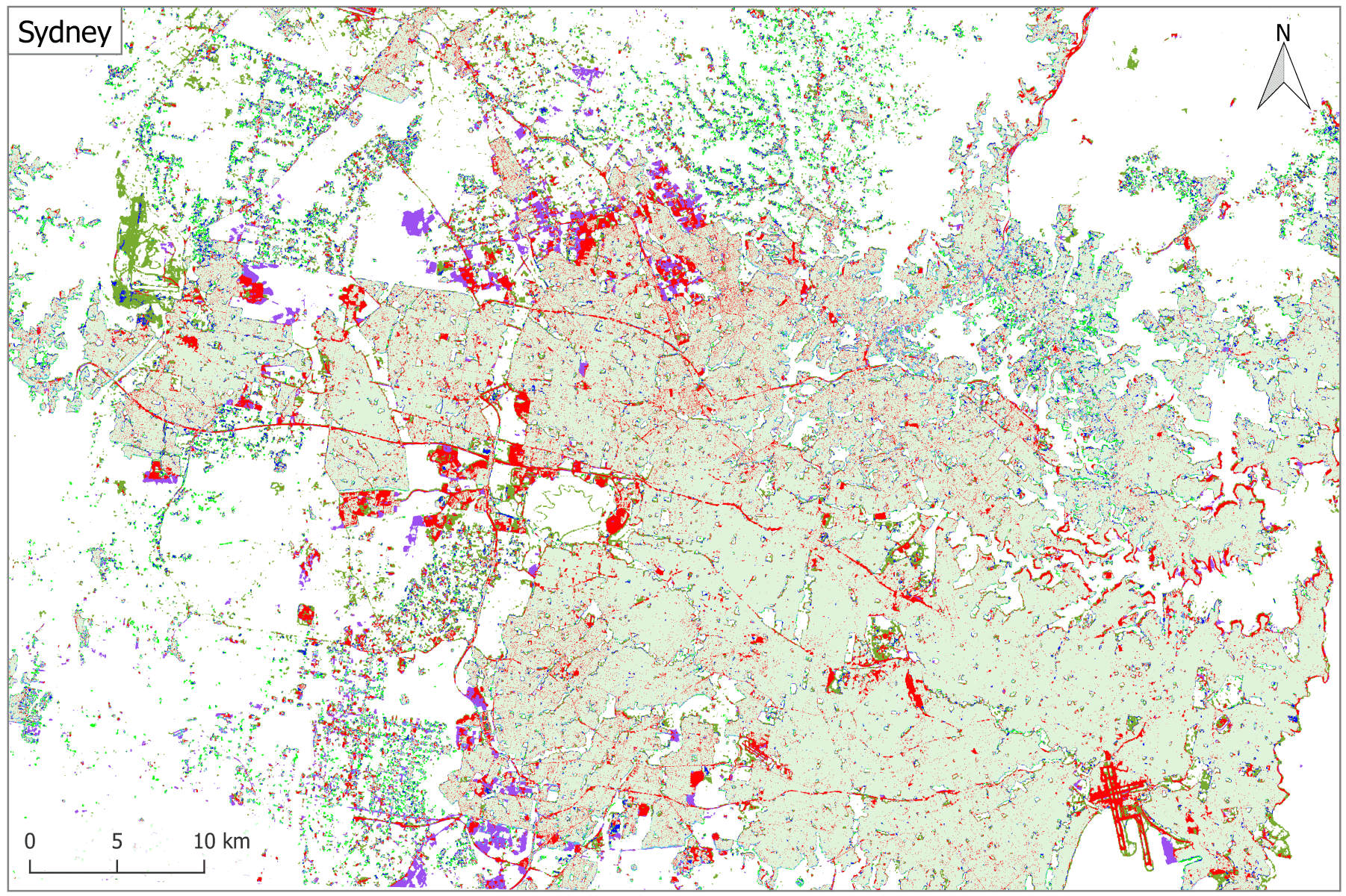}
\caption{Continued.}
\label{fig:cmp_diff}
\end{figure*}

\begin{table}[!tbh]
  \centering
  \scriptsize
  \caption{\textcolor{black}{Interpretation of the colors in Fig. \ref{fig:cmp_diff}. FP and FN are false positive and false negative, i.e., commission error and omission error, respectively.}}
    \begin{tabular}{cccccccccc}
    \toprule
    \multicolumn{2}{c}{} & \multicolumn{2}{c}{Our results} &       & \multicolumn{2}{c}{GUF} &       & \multicolumn{2}{c}{GHSL} \\
    \midrule
          &       & \cellcolor{p}     & \cellcolor{gugs}  &       & \cellcolor{gu} & \cellcolor{gsp}   &       & \cellcolor{gh} & \cellcolor{gup} \\
\cmidrule{3-4}\cmidrule{6-7}\cmidrule{9-10}    \multirow{2}[2]{*}{mapping
target?} & yes   & correct & FN    &       & correct & FN    &       & correct & FN \\
          & no    & FP    & correct &       & FP    & correct &       & FP    & correct \\
    \bottomrule
    \end{tabular}%
  \label{tab:interpretationColor}%
\end{table}

\begin{figure}[!tbh]
	\centering
    \includegraphics[width=0.8\textwidth]{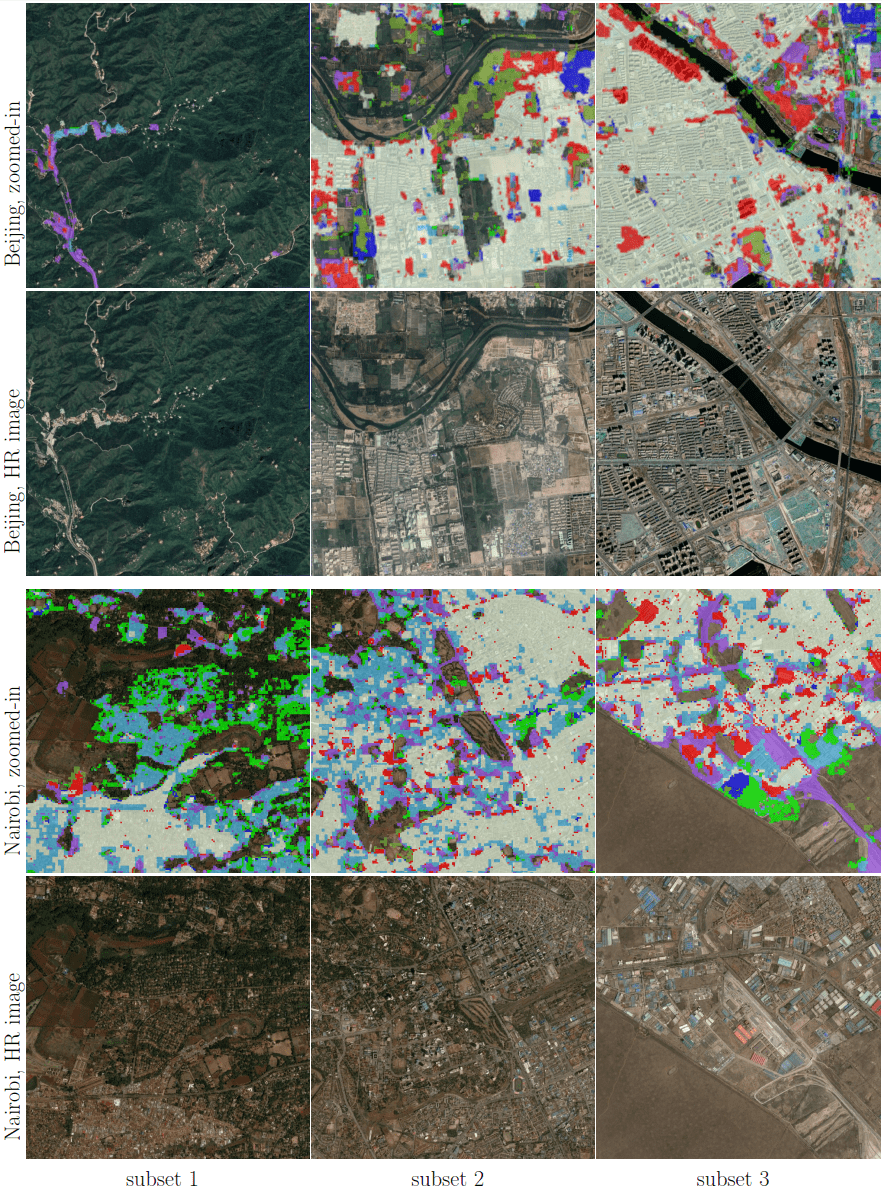}
	\caption{{Closer view of the three subsets of sample test scenes distributed across the world, overlaid on  high resolution images. The high resolution images are also shown for detailed interpretation.}
}
	\label{fig:cmp_diff_subset}
\end{figure}
\begin{figure}[!htb]
\centering
    \ContinuedFloat
\includegraphics[width=0.8\textwidth]{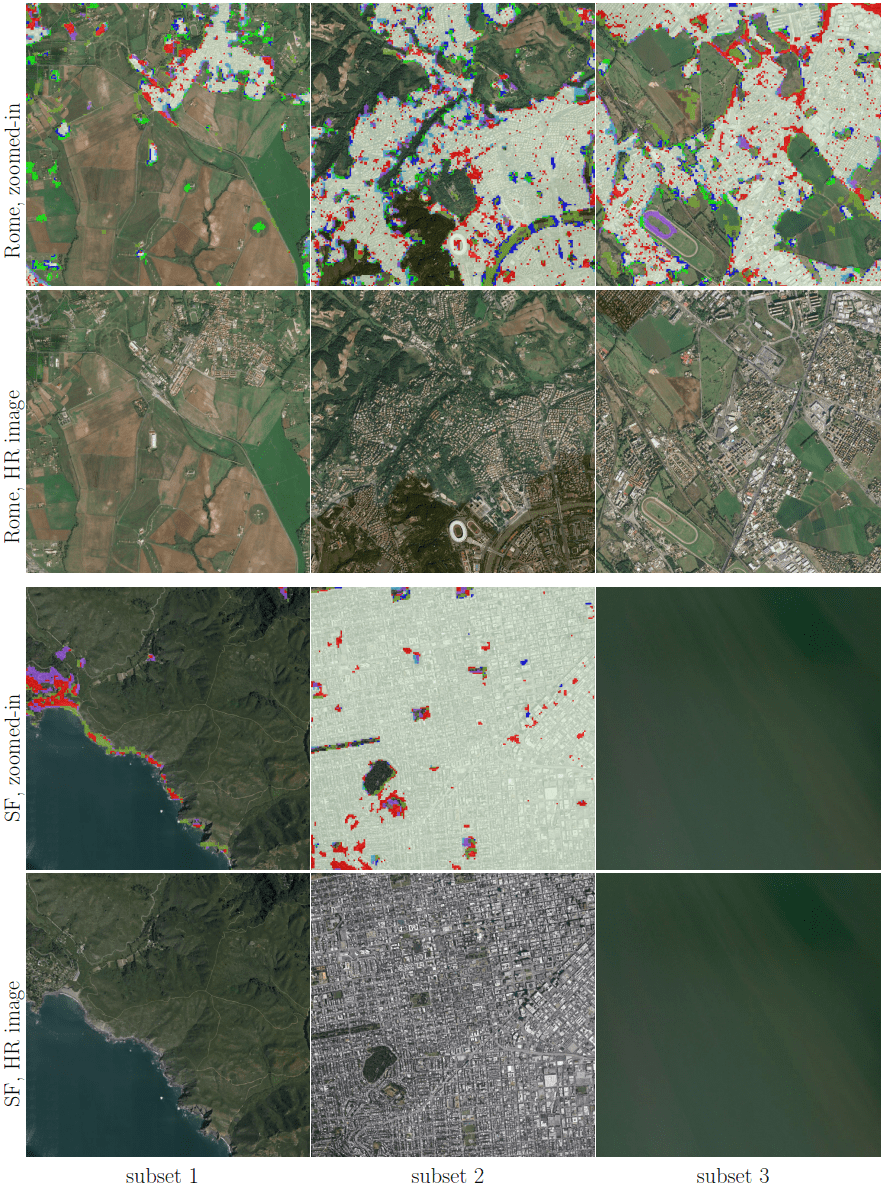}
    \caption{Continued.}
\end{figure}
\begin{figure}[!htb]
\centering
    \ContinuedFloat
\includegraphics[width=0.8\textwidth]{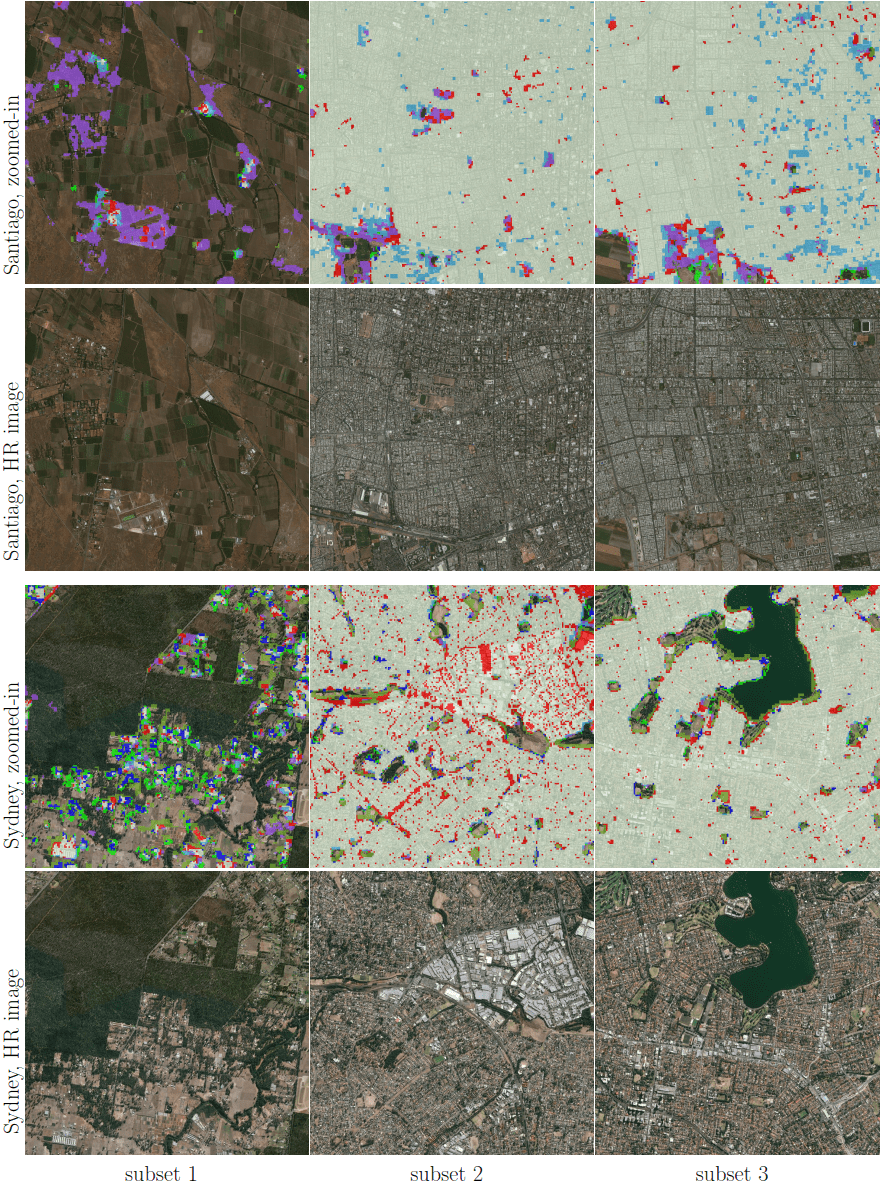}
    \caption{Continued.}
\end{figure}

\subsection{Examples of regional-scale and country-wide HSE mapping}
\label{exampleMapping}
\begin{figure}[!tbh]
	\centering
	\includegraphics[width=0.99\textwidth]{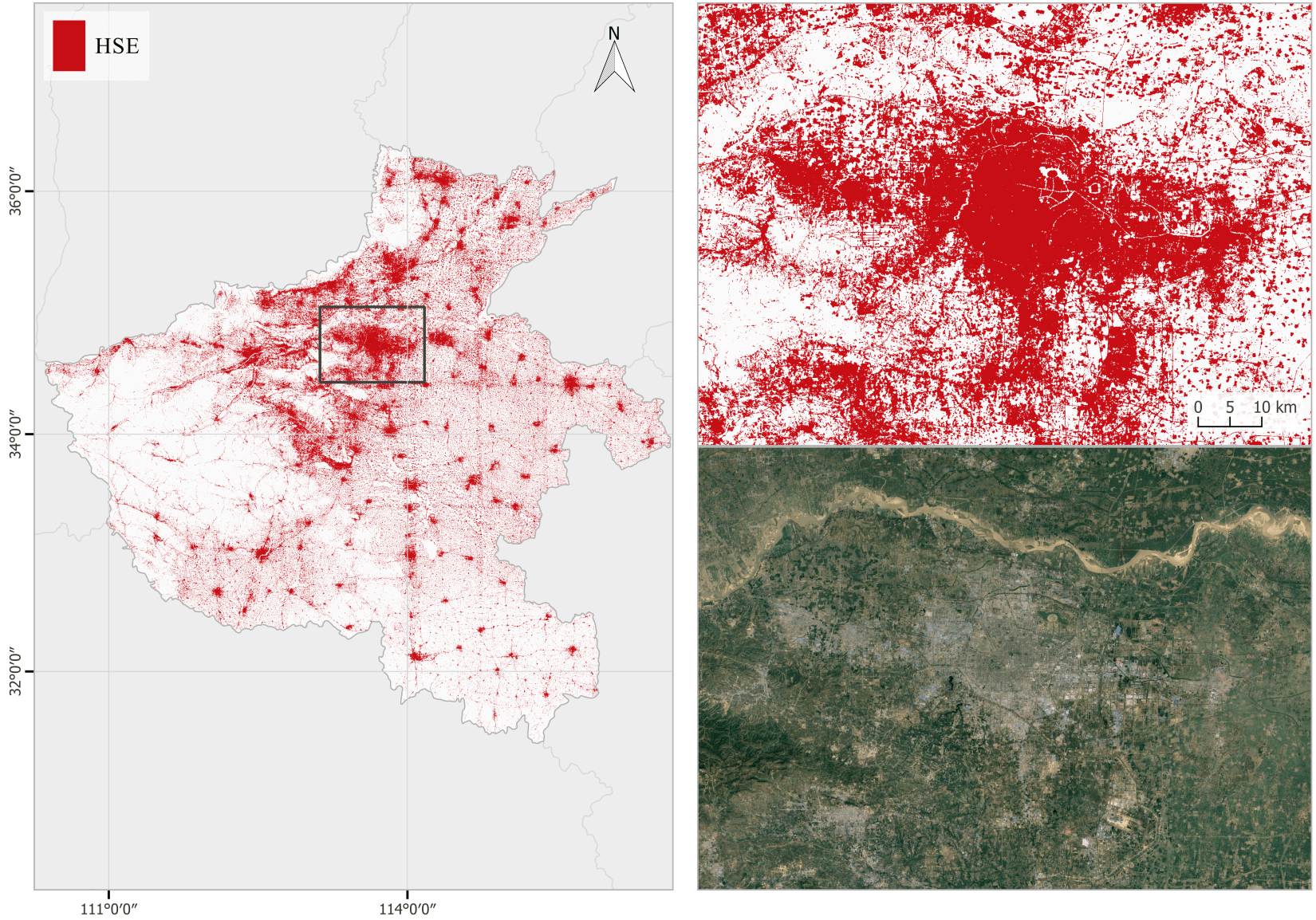}
	\caption{{Regional HSE mapping example in Henan (province), China. The Zhengzhou (city) area is zoomed in and compared to a high resolution image.}}
	\label{fig:hn}
\end{figure}

\begin{figure}[!tbh]
	\centering
	\includegraphics[width=0.99\textwidth]{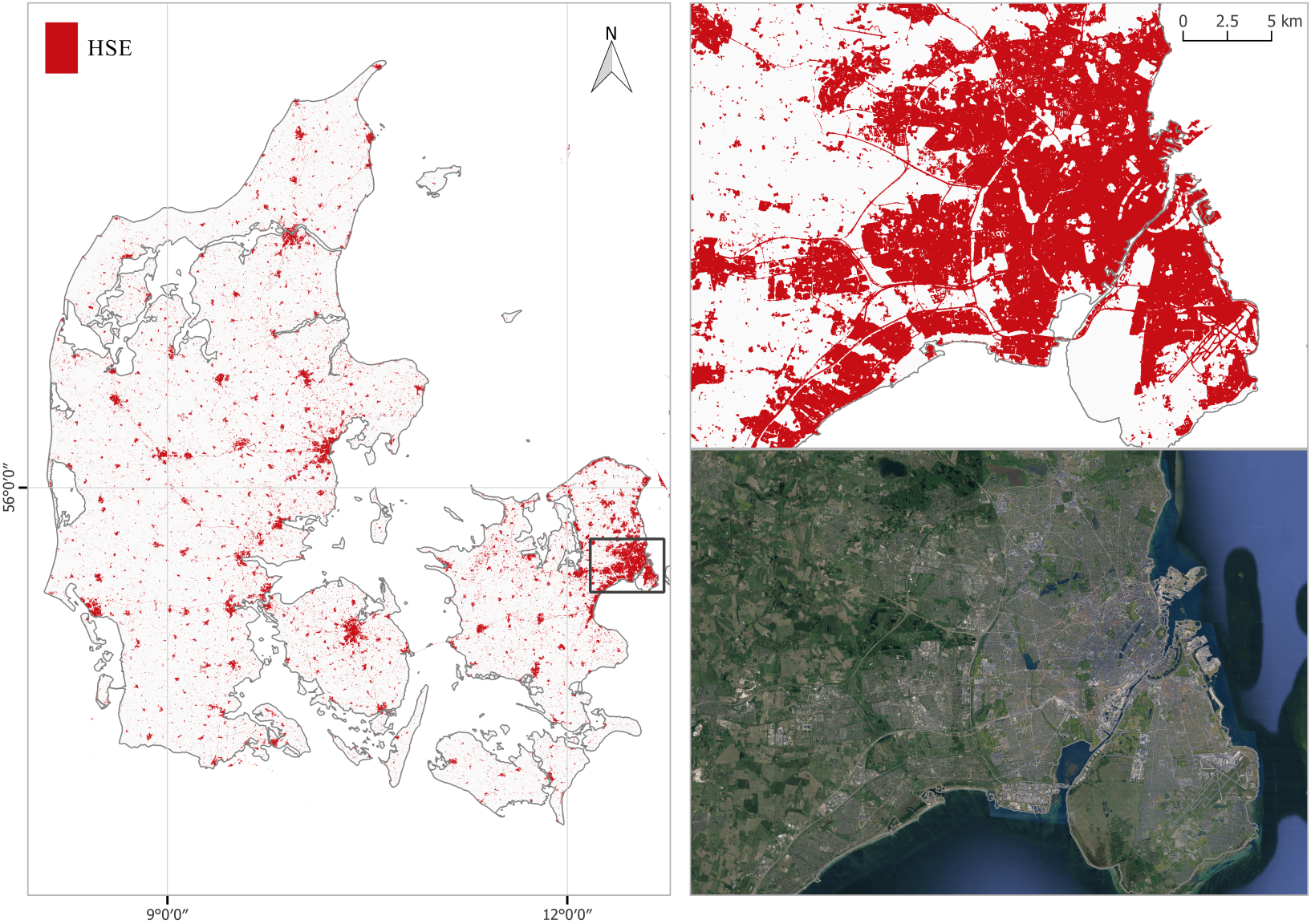}
	\caption{{Country-wide HSE mapping example in Denmark. The Copenhagen area is zoomed in and compared to a high resolution image.}}
	\label{fig:dk}
\end{figure}

In order to validate the stability of the proposed framework, we tested the workflow on a regional-scale and country-wide HSE mapping task, in Henan province, China and in Denmark. The total area of each is about 167,000 and 42,933 $km^2$ \textcolor{black}{, respectively}. The HSE mapping results are shown in Figs. \ref{fig:hn} and \ref{fig:dk}. The general urban pattern is successfully mapped for both examples, as can be seen when they are compared with high resolutions satellite images. This test demonstrates the general performance and the potential for upscaling of the presented framework.

\section{Discussion}
\label{sec:dis}

In the section, we provide \textcolor{black}{some empirical evidence of the framework setup and network design, as well as} addressing the problems and questions posed in Sec. \ref{sec:intro}, using insights gained from the extensive experimental results presented in Sec. \ref{sec:res} and some additional investigations. This section will also discuss some lessons learned that are relevant to similar topics and further possible improvements toward more accurate and operational HSE mapping.

\subsection{\textcolor{black}{Choice of the proposed framework}}
\textcolor{black}{To demonstrate the rationale behind our design choice, this section provides some \textcolor{black}{sensitivity analyses}. The achieved results are first compared to those from several state-of-the-art baseline methods in Sec. \ref{sec:cmpBaselinesM}. In addition, two different ways of splitting of training and validation data are compared to justify our experimental setup. Last, the effect of network depth and width are investigated for the employed architecture, to provide more insights into our approach.}

\subsubsection{\textcolor{black}{Comparison with baseline methods}}
\label{sec:cmpBaselinesM}

\textcolor{black}{The achieved HSE mapping results from the proposed Sen2HSE-Net are compared to those from baseline networks in Tab.  \ref{tab:sotaNN} for test, both beyond and within Europe. Table  \ref{tab:sotaNN} shows that the proposed shallow network with 9 layers is able to provide even better mapping accuracy than the much deeper and relatively complicated U-Net \cite{ronneberger2015u}, with more trainable parameters. In addition, the achieved results from Sen2HSE-Net are much more accurate than those from ResNet-PSPNet \cite{zhao2017pyramid}, ResNet-FCN-8 \cite{long2015fully}, and attention-based FCN \cite{fu2018dual}, which have been shown to be more powerful for detailed semantic segmentation. One possible reason is the information loss from the pooling layers in the encoding process (by ResNet), which is not suitable for our HSE mapping task and Sentinel-2 data. Furthermore, this loss cannot be compensated, even with the sophisticated design of the decoding part, either with pyramid scene parsing by ResNet-PSPNet, or upsampling with low-level features considered by ResNet-FCN-8, or attention modules proposed in \cite{fu2018dual}. These observations confirm the assumptions that motivate our framework design: good performance is not guaranteed when simply and directly using the state-of-the-art networks for remote sensing tasks. Instead, characteristics of both the task and data need to be integrated into the network design. Additionally, Tab. \ref{tab:sotaNN} shows that it is possible to use a simple FCN to achieve promising HSE mapping results, instead of relying on the existing rather sophisticated networks. Even though comparable results can be achieved from directing employing U-Net, the proposed Sen2HSE-Net is much lighter, which is significant for large-scale mapping.}

\begin{table}[H]
  \centering
  \scriptsize
  \caption{Results from Sen2HSE-Net and three baseline semantic segmentation networks, tested in areas beyond and within Europe.}
    \begin{tabular}{ccccccccccccc}
    \toprule
    \multirow{2}[4]{*}{Method} & \multicolumn{4}{c}{test beyond Europe } &       & \multicolumn{4}{c}{test in Europe} &       & \multicolumn{2}{c}{network} \\
\cmidrule{2-5}\cmidrule{7-10}\cmidrule{12-13}          & Kappa & AA    & recall & F1    &       & Kappa & AA    & recall & F1    &       & \multicolumn{1}{l}{layer} & \multicolumn{1}{l}{\# of Para.} \\
    \midrule
    Sen2HSE-Net & 0.809 & 90.1\% & 91.4\% & 0.906 &       & 0.802 & 90.5\% & 84.4\% & 0.834 &       & 9     & 1,124,866 \\
    U-Net \cite{ronneberger2015u} & 0.804 & 90.3\% & 90.4\% & 0.903 &       & 0.788 & 89.3\% & 81.7\% & 0.822 &       & 24    & 31,036,872 \\
    ResNet-PSPNet \cite{zhao2017pyramid} & 0.655 & 82.8\% & 80.3\% & 0.824 &       & 0.644 & 80.3\% & 64.5\% & 0.697 &       & 58    & 28,550,594 \\
    ResNet-FCN-8 \cite{long2015fully} & 0.740 & 87.2\% & 89.3\% & 0.875 &       & 0.719 & 84.8\% & 73.0\% & 0.762 &       & 60    & 31,960,710 \\
    FCN+dual attention \cite{fu2018dual}& 0.785 & 89.4\% & 86.3\% & 0.888 &       & 0.760 & 85.2\% & 72.4\% & 0.795 &       & 27    & 14,405,056 \\
    \bottomrule
    \end{tabular}%
  \label{tab:sotaNN}%
\end{table}%

\subsubsection{\textcolor{black}{Effect of training and validation data split}}
\textcolor{black}{Testing performance depends on how the training and validation datasets are split, because validation data provides hints of the progress during training and is the basis for choosing the best trained model. To understand the influence of the choice of validation data on the eventual test results, we have investigated two different variants of validation data selection. It has to be noted that the validation data is always chosen as a subset of the training set from the training scenes, whereas the test data in this study always came from test scenes and remained unseen during training.}

\textcolor{black}{This effect is shown in Tab. \ref{tab:setupNN}, with both the proposed Sen2HSE-Net and the standard segmentation network, U-Net, as examples. Random split is randomly choosing about 25\% of the data from each training city as the validation dataset, while spatial split is extracting about 25\% of the upper left part of each city as the validation dataset. Spatial split is also what is used in this study. As described in Tab. \ref{tab:cityChrac}, the models are tested on test data that are completely unseen during training. From the illustration in Tab. \ref{tab:setupNN}, we can see that spatial split is better than random split, since almost all metrics are better from a spatial split, which is true for both networks. This may be because the distribution of training and validation data is more similar in random split than spatial split, which leads to a validation accuracy that is closer to the training accuracy. As a result, the chosen model is optimal for the training areas, rather than the unseen test areas.}

\begin{table}[H]
  \centering
  \scriptsize  
  \caption{Results from different approaches to splitting of training and validation datasets, tested in \textcolor{black}{completely unseen} areas both beyond and within Europe.}
    \begin{tabular}{ccccccccccc}
    \toprule
    \multicolumn{2}{c}{\multirow{2}[3]{*}{network and data split}} & \multicolumn{4}{c}{test beyond Europe } &       & \multicolumn{4}{c}{test in Europe} \\
\cmidrule{3-6}\cmidrule{8-11}    \multicolumn{2}{c}{} & Kappa & AA    & recall & F1    &       & Kappa & AA    & recall & F1 \\
 \midrule
    \multirow{2}[3]{*}{Sen2HSE-Net} & spatial & 0.809 & 90.1\% & 91.4\% & 0.906 &       & 0.802 & 90.5\% & 84.4\% & 0.834 \\
\cmidrule{2-11}          & random & 0.788 & 89.1\% & 87.7\% & 0.891 &       & 0.798 & 89.6\% & 82.1\% & 0.830 \\
    \midrule
    \multirow{2}[4]{*}{U-Net} & spatial & 0.806 & 90.0\% & 90.2\% & 0.902 &       & 0.801 & 89.2\% & 81.1\% & 0.832 \\
\cmidrule{2-11}          & random & 0.805 & 90.1\% & 87.7\% & 0.897 &       & 0.791 & 88.5\% & 79.5\% & 0.824 \\
    \bottomrule
    \end{tabular}%
  \label{tab:setupNN}%
\end{table}%

\subsubsection{\textcolor{black}{Effect of network depth and width}}

\textcolor{black}{It is important to know whether better HSE mapping results can be achieved from a deeper and wider version of Sen2HSE-Net using the setup of this study.
Table \ref{tab:shapeNN} sheds light on this potential improvement, by comparing the results of one wider and three deeper versions, as well as the number of trainable parameters in each network. From these comparisons, we observe no gain from a wider network and a slight improvement from a deeper network. Interestingly, the improvement is not present when the depth increases further, from depth 13 to 17 and 21. This might result from the characteristics of the task, \textcolor{black}{the use of Sentinel-2} images, which are not high resolution, as well as the testing choice (in unseen areas).}

\begin{table}[H]
  \centering
    \scriptsize
  \caption{Results from Sen2HSE-Net of varying depth and width. All comparing networks employ the same overall architecture as Fig. \ref{fig:network}. The result in the first row is from the configuration used in Sec. \ref{sec:res}.}
    \begin{tabular}{ccccccccccccc}
    \toprule
    \multicolumn{3}{c}{network} &       & \multicolumn{4}{c}{test beyond Europe } &       & \multicolumn{4}{c}{test in Europe} \\
\cmidrule{1-3}\cmidrule{5-8}\cmidrule{10-13}    \multicolumn{1}{l}{\# of first Conv} & \multicolumn{1}{l}{layer} & \multicolumn{1}{l}{\# of Para.} &       & Kappa & AA    & recall & F1    &       & Kappa & AA    & recall & F1 \\
\cmidrule{1-3}\cmidrule{5-13}    \multirow{4}[4]{*}{f=16} & 2+2+2+2+1 & 1,124,866 &       & 0.81  & 90.1\% & 91.4\% & 0.91  &       & 0.80  & 90.5\% & 84.4\% & 0.83 \\
\cmidrule{2-3}\cmidrule{5-13}          & 3+3+3+3+1 & 1,874,098 &       & 0.82  & 90.6\% & 93.1\% & 0.91  &       & 0.80  & 90.4\% & 84.2\% & 0.83 \\
          & 4+4+4+4+1 & 2,623,330 &       & 0.81  & 90.1\% & 91.5\% & 0.90  &       & 0.80  & 90.7\% & 85.0\% & 0.84 \\
          & 5+5+5+5+1 & 3,372,562 &       & 0.80  & 89.7\% & 92.5\% & 0.90  &       & 0.80  & 90.3\% & 84.2\% & 0.83 \\
\cmidrule{1-3}\cmidrule{5-13}    f=32  & 2+2+2+2+1 & 4,493,826 &       & 0.81  & 90.1\% & 90.0\% & 0.90  &       & 0.80  & 89.0\% & 80.7\% & 0.83 \\
    \bottomrule
    \end{tabular}%
  \label{tab:shapeNN}%
\end{table}%

\subsection{\textcolor{black}{Analysis} of the HSE mapping framework}
\textcolor{black}{While the quantitative and qualitative results presented in Sec. \ref{sec:res} have shown the promising performance of our framework, there are some details requiring analysis for a better understanding of both the method and the produced results. These details will be addressed in this subsection.}

\subsubsection{\textcolor{black}{Mapping power of the proposed framework}}

\begin{figure}[!tbh]
	\centering
    \includegraphics[width=0.95\textwidth]{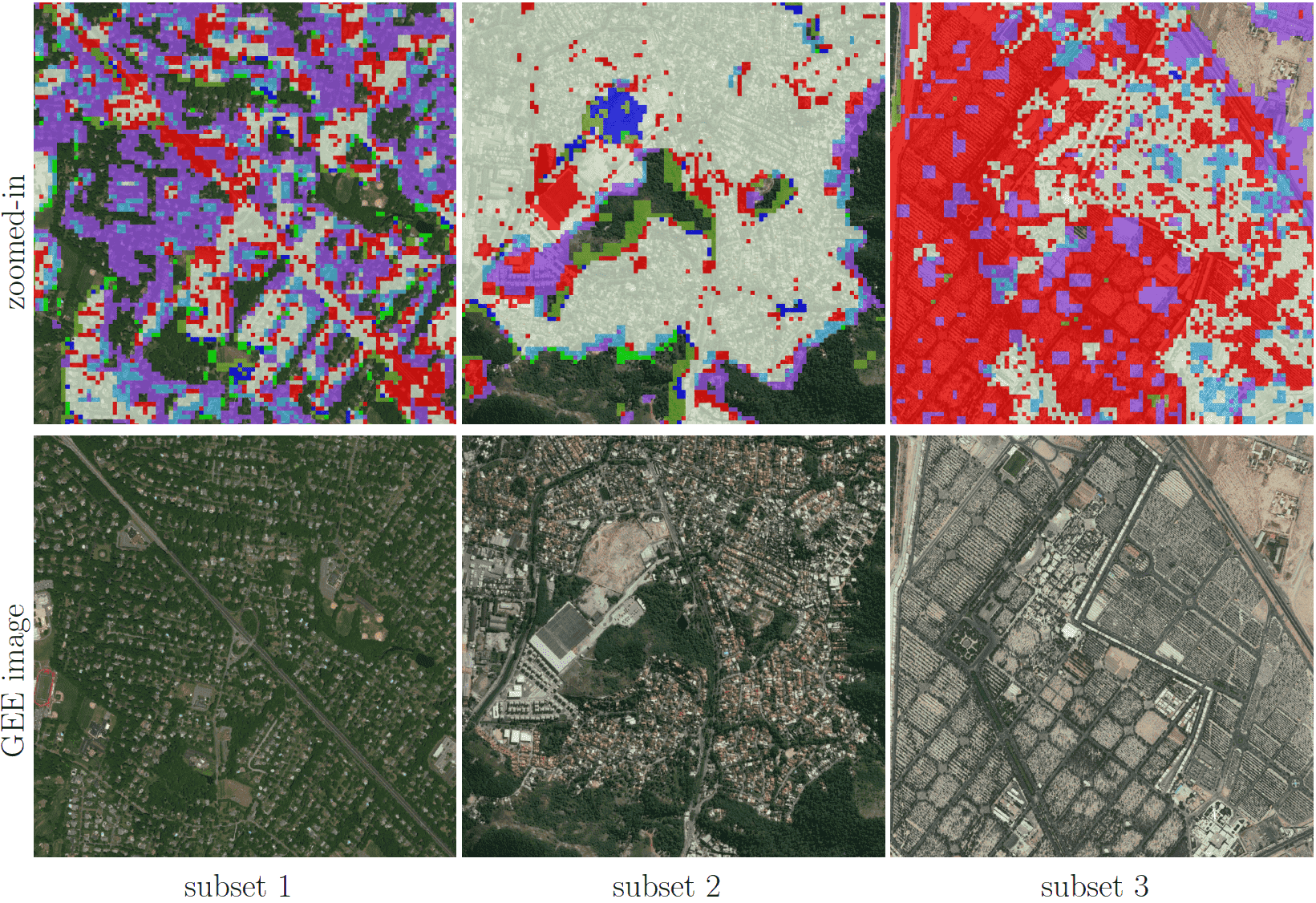}
	\caption{{Closer view of some positive examples, with the same legend as in Fig. \ref{fig:cmp_diff}. Colors can be interpreted according to Tab. \ref{tab:interpretationColor}.}
}
	\label{fig:cmp_diff_strong}
\end{figure}

\textcolor{black}{The goal of this study is to explore a better solution for mapping HSE with the potential of upscaling. Figure \ref{fig:cmp_diff_strong} illustrates the HSE mapping power, with some positive examples in test scenes in New York City, Rio, and Tehran. In the examples in Fig. \ref{fig:cmp_diff_strong}, only our solution is able to include sparse buildings and buildings on the boundaries, surrounded by trees and gardens, as the purple outlines indicate the areas that are only mapped by our results and are missed by the other two baseline products. This can also be observed in the first subset in Beijing, the first subset in Nairobi, and the first subset in Santiago, as shown in Fig. \ref{fig:cmp_diff_subset}. In the second subset of Fig. \ref{fig:cmp_diff_strong}, only our result is able to exclude the soil ground from the mapping results, as the blue outlines indicate areas mapped by other layers but not by our results. This can also be seen in the second subset of Beijing and the third subset of Nairobi, as shown in Fig. \ref{fig:cmp_diff_subset}. The red and cyan color outlines indicate areas that are not mapped by GUF and GHSL, respectively. Since these areas are mapped not only by our results but also by one of the baseline datasets, they are very likely HSE, and correctly detected by our approach.  This can be seen from the third subset in Fig. \ref{fig:cmp_diff_strong}, as well as the first subset in Nairobi, the first subset in Rome, and the second subset in Sydney in Fig. \ref{fig:cmp_diff_subset}.}

\begin{figure}[!tbh]
	\centering
    \includegraphics[width=0.99\textwidth]{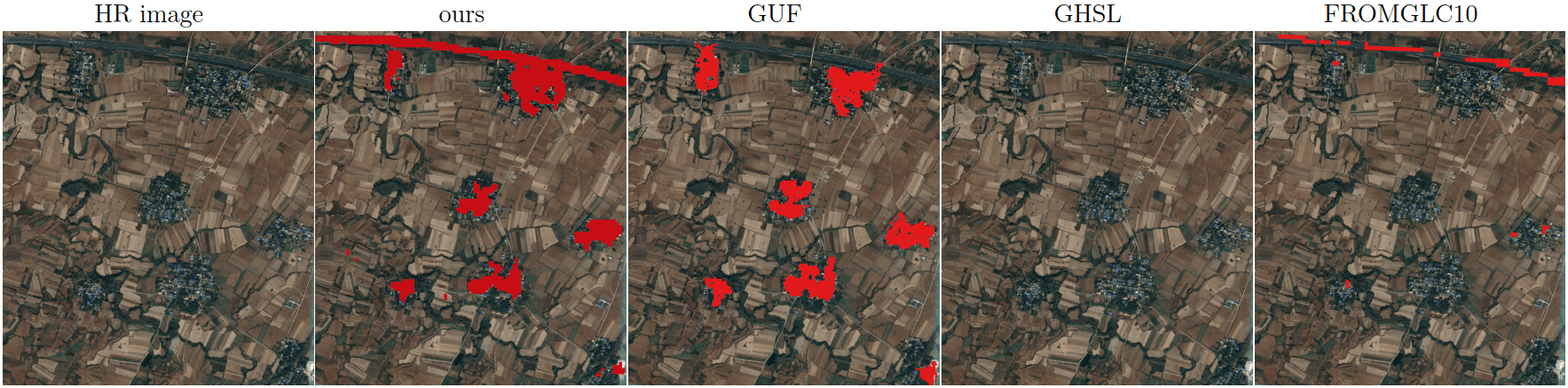}
	\caption{{Closer view of the HSE mapping power of the proposed framework, with an example around the location of longitude 113.2072 and latitude 32.6849.}
}
	\label{fig:cmp_home}
\end{figure}

\textcolor{black}{More evidence of the mapping power of the proposed framework can be seen in Fig. \ref{fig:cmp_home}, a close-up of Fig. \ref{fig:hn}, where we are able to detect buildings in small villages as well as GUF, which is derived from very high resolution SAR images. The other products unfortunately fail to map these areas. This also shows the improvement of space over current land cover mapping at global scale, especially in rural areas.}

\textcolor{black}{Jointly considering the accuracy assessment with respect to the MLGCPs and the OSM building layer shown in Tab. \ref{tab:MLGCPs_osm_cmp}, as well as the visualizations at different scales in Figs. \ref{fig:cmp_diff} and \ref{fig:cmp_diff_subset}}, we conclude that HSE maps can be created by the proposed approach, with comparable or even better quality than state-of-the-art products. Generally good results can be achieved, \textcolor{black}{even in} test cities with various typologies of urban areas and vegetation, different climate\textcolor{black}{,} and \textcolor{black}{diverse} culture regions. This finding suggests the proposed framework's potential for generalizing and upscaling. Furthermore, the assessment of the experimental results provides evidence that the motivation for setting up the framework is valid. That is simple FCNs and the multi-spectral images from the Sentinel-2 mission are indeed valuable for large-scale HSE mapping and could be exploited to produce large-scale HSE maps with a 20 m GSD. Also, \textcolor{black}{this work demonstrates that} not having highly accurate pixel-level ground truth does not hinder the successful adaptation of deep neural networks \textcolor{black}{to} the application of HSE mapping.

\begin{figure}[!tbh]
	\centering
    \includegraphics[width=0.95\textwidth]{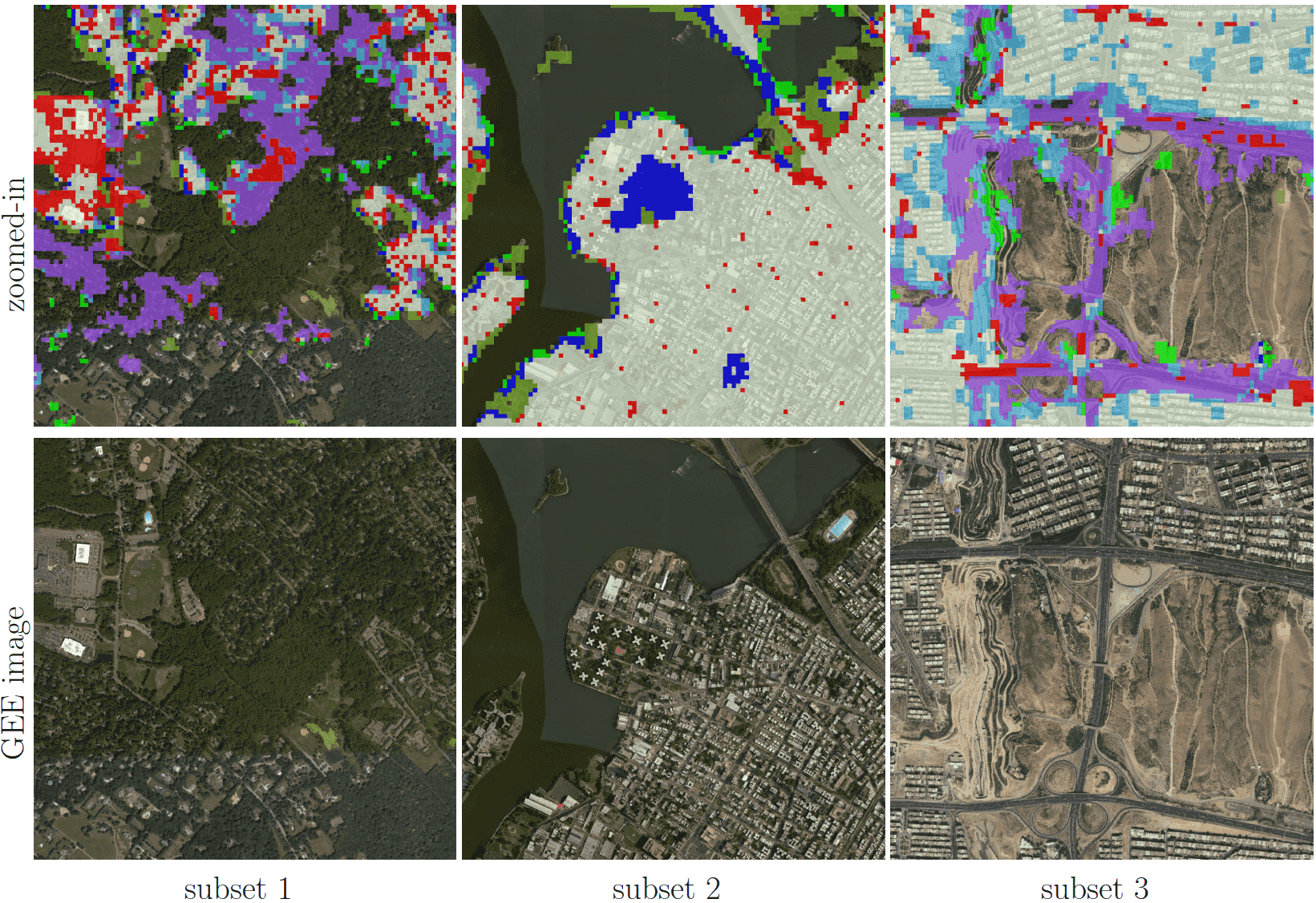}
	\caption{{Closer view of some negative examples, with the same legend as in Fig. \ref{fig:cmp_diff}. Colors can be interpreted according to Tab. \ref{tab:interpretationColor}.}
}
	\label{fig:cmp_diff_weak}
\end{figure}

\textcolor{black}{However, some problems in the current mapping results remain, as shown by the negative examples from test scenes in New York City and Tehran in Fig. \ref{fig:cmp_diff_weak}. In the first subset, there are still some buildings omitted by our mapping results, and in the second subset, there is still an area omitted only by our approach. In addition, some overestimation can be seen in the third subset; this can also be observed in the first subset of Rome and the third subset of Santiago in Fig. \ref{fig:cmp_diff_subset}. This overestimation, i.e., a commission error, is inherent to the definition of the task and the setup of the framework. Specifically, the goal is to detect whether there is HSE in a 20 by 20 meter cell. Therefore, the boundaries tend to be identified as HSE. Possible approaches for improvement will be proposed in Sec. \ref{sec:furtherImp}.}

\subsubsection{Differences between HSE mapping results and baseline products}

\begin{figure}[!tbh]
	\centering
	\includegraphics[width=1\textwidth]{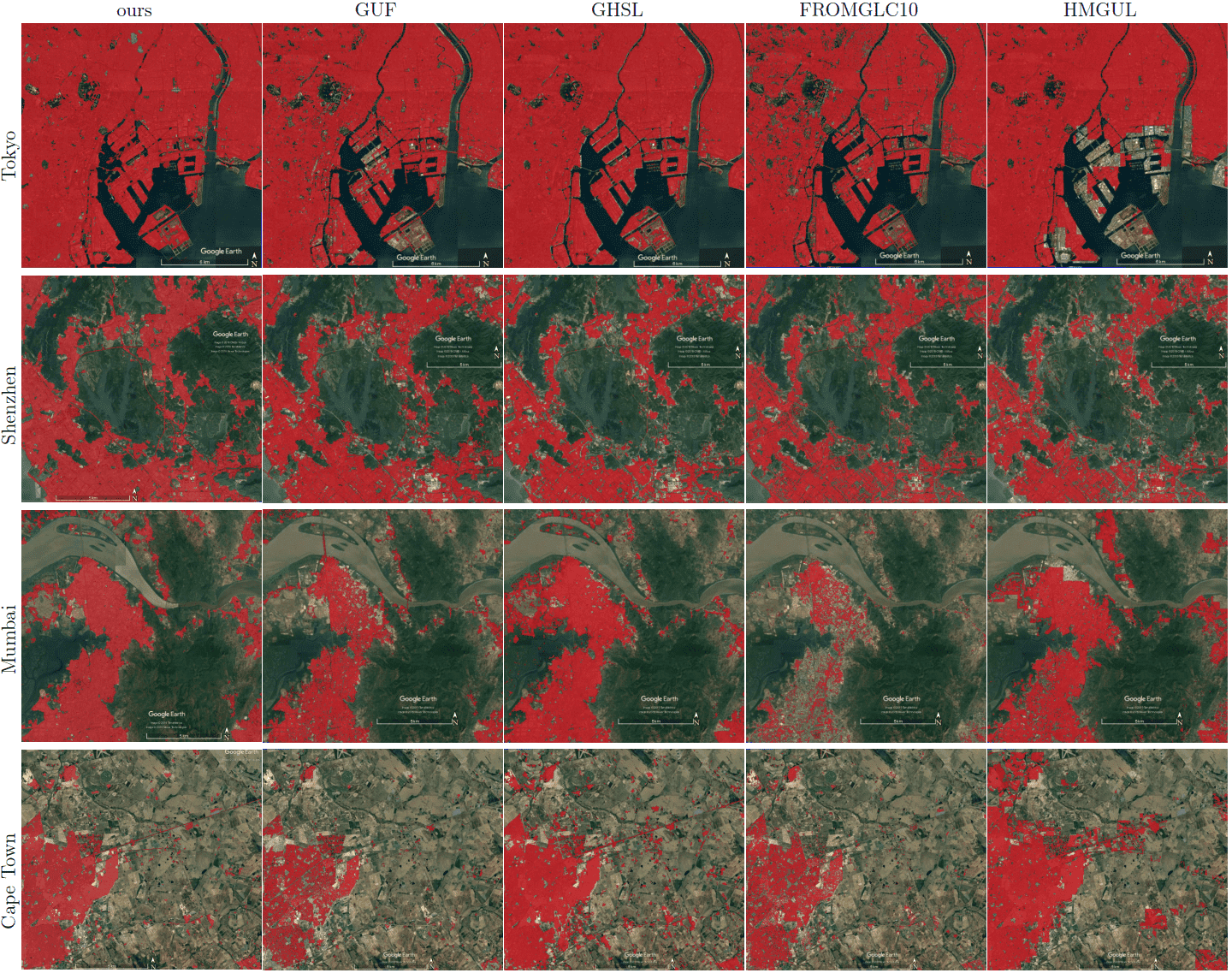}
	\caption{{Differences among HSE-related datasets. Red areas are mapped areas from existing products based on Tab. \ref{tab:refData}. Four distinct areas are chosen to present highly heterogeneous urban structures from different parts of the world.}}
	\label{fig:diff_datasets}
\end{figure}

Comparisons in Sec. \ref{sec:res} also reveal some notable differences among our HSE mapping results, GHS built-up grid, GUF, FROM-GLC10, and HMGUL. These differences are further visualized in Fig. \ref{fig:diff_datasets} for four distinct areas around the world.
Similar to the HSE mapped by our approach, both GHSL and HMGUL include not only buildings but also impervious surfaces such as roads and parking lots, even though they are not focused on impervious surfaces. This is because the medium-resolution data \textcolor{black}{employed} is not enough to exclude small gaps among buildings, especially when the gaps are covered by the same materials as buildings. \textcolor{black}{It} is \textcolor{black}{thus} challenging to distinguish these areas that are highly related to HSE and bear a similar spectral signature as buildings \textcolor{black}{when} using the spectral information from optical satellite images. \textcolor{black}{In contrast,} GUF does not contain such impervious surfaces, as can be seen from the red regions in Figs. \ref{fig:cmp_diff} and \ref{fig:cmp_diff_subset}. This is because GUF focuses more on vertical building structures, \textcolor{black}{removing} roads and paved surfaces during the post-editing period \cite{esch2017breaking}. It is also \textcolor{black}{due to} the peculiarities of the SAR images used for the production of GUF. The local speckle information and the texture information in the SAR images makes it possible to specifically detect vertical structures such as buildings \cite{klotz2016good, qiu2018towards}. Specifically, buildings are characterized by stronger back-scattering signals than airport roads, even though they are made of the same materials. However, when using optical satellite images, it is challenging to distinguish different land covers within the super-class of impervious surfaces, as they share similar spectral signatures. An illustrative example is the \textcolor{black}{Sydney Airport} (the red cross-shape in the lower right corner of Fig. \ref{fig:cmp_diff}), where the aircraft runways are mapped as built-up areas in both GHSL and the result of this study. 

\textcolor{black}{Also due to the peculiarities of the SAR images used for the production of GUF, some sparse trees can be mistaken as buildings, as shown in the the first and third subset of Nairobi in Fig. \ref{fig:cmp_diff_subset}. For GHSL, error prone areas are forests and bodies of water, as shown in the second subset of Rome, the first subset of San Francisco, and the third subset of Sydney. As a result of the two phenomenon discussed above, sparsely built-up areas surrounded by sparse forest can be challenging, as can be seen from the ``noisy'' visualizations in the suburban areas in Fig. \ref{fig:cmp_diff}.}

In Fig. \ref{fig:diff_datasets}, it can be \textcolor{black}{seen} that FROM-GLC10 and HMGUL are subject to obvious omission error\textcolor{black}{s} in Mumbai and Tokyo, respectively, providing one more \textcolor{black}{piece of} evidence for the proposed approach's improved performance over state-of-the-art layers. A further comparison between our results and GHSL shows that more roads are mapped by our approach, as shown by the purple lines in the Nairobi and Beijing test scenes in Fig. \ref{fig:cmp_diff}, demonstrating the powerful mapping \textcolor{black}{capability} of our framework.

These characteristics of each product \textcolor{black}{discussed above} relate to the differing definitions of ``urban\textcolor{black}{,}'' ``human settlement\textcolor{black}{,}'' \textcolor{black}{and} ``built-up,'' as well as the the mapping approaches employed and the datasets used. End users of these products in particular should take note of these differences. On the other hand, understanding these differing characteristics also makes \textcolor{black}{it} possible to extract complementary information from different products for various applications. 

It should be mentioned that all comparisons in this study are \textcolor{black}{intended} merely to provide an assessment with reference to the state-of-the-art products. The \textcolor{black}{occasional} inferior performance of the GHS built-up grid and GUF is certainly partially {due to temporal gaps in data collection:} the ongoing urbanization of the world has changed many originally suburban areas to newly built-up areas after the GHS built-up grid and GUF \textcolor{black}{were} released. This \textcolor{black}{cannot} be easily ignored, especially for cities in developing countries\textcolor{black}{,} such as Beijing. This issue highlights the necessity for up-to-date worldwide HSE information, in addition to the existing products\textcolor{black}{:} GUF\textcolor{black}{,} with \textcolor{black}{its} unprecedented spatial resolutions, the GHS built-up grid, with \textcolor{black}{its} multi-temporal resolution {, and FROM-GLC10\textcolor{black}{,} with \textcolor{black}{its} detailed land cover information}.

\subsection{Further improvements toward operational mapping}
\label{sec:furtherImp}
We are able to achieve state-of-the-art HSE results for several representative scenes across the world. Furthermore, comparable accuracy is achieved for both regional mapping (\textcolor{black}{three test scenes in Europe}) and large-scale mapping (the ten world-wide distributed test cities), \textcolor{black}{as shown in Tab. \ref{tab:sotaNN}}. However, there is still much room for further improvements toward an operational {large-scale---even global---}process. The improvements can mainly be achieved \textcolor{black}{with respect to} three aspects: the input satellite images, the deep neural network architectures, and the post-processing of the mapped HSE results. {First, Level-2A Sentinel-2 images (bottom-of-atmosphere reflectance) and the spectral ratios could bring accuracy improvement.} In order to produce HSE maps at a regular frequency, it is not enough using Sentinel-2 images alone, especially in regions with heavy cloud cover throughout the year such as the Southeast Asia \cite{stengel2017cloud}. One solution is to \textcolor{black}{employ} multi-sensor, multi-temporal, and multi-modal data fusion, thus improving accuracy and enhancing temporal and spatial sampling \cite{schmitt2016data, ghamisi2018multisource, lefebvre2016monitoring, hong2019cospace, hong2019learnable, hong2019augmented, qiu2019fusingLetter}. Considering the scale and aiming applications, Landsat-8 and Sentinel-1 images could also be exploited for HSE mapping. It should be mentioned that the proposed framework can be easily adapted for these two datasets \textcolor{black}{after proper preprocessing, like filtering for SAR images and cloud removal for optical images}. In addition to the input images, improvement can also be realized via an ensemble with other deep CNNs in order to take advantage of their complementary characteristics and heterogeneous properties, as demonstrated by \cite{noh2015learning}. Furthermore, the performance of the proposed framework should be further investigated and evaluated in rural areas, where built-up areas tend to be sparse and can be easily omitted. Finally, once the HSE results are acquired, further post-processing could be carried out independently for each city. For instance, a conditional random field could be applied to the output mapping results, in order to homogenize the segmentation \textcolor{black}{\citep{maggiolo2018improving}}. Furthermore, in this process, any locally available datasets such as census data, as well as prior knowledge, could be exploited. Other directions worth exploring include adapting the trained model with semi-supervised learning-based strategies and transfer learning, including multitask learning, domain generalization, and domain adaptation, for the purpose of better generalization \cite{tuia2016domain}.


\section{Conclusions and outlook}
\label{sec:conc}

Detailed and up-to-date HSE maps provide essential information about the human footprint on the earth, 
thus making sustainable development possible via proactive conservation. This paper presents a framework for large-scale HSE mapping from Sentinel-2 images, by exploiting \textcolor[rgb]{0,0,0}{a shallow yet effective} FCN for semantic segmentation. In particular, the newly proposed framework 
 takes advantages of \textcolor[rgb]{0,0,0}{globally available} images from the Sentinel-2 mission, {featuring medium spatial resolution}, high revisit time, and multi-spectral imaging. As demonstrated in this paper, higher accuracy than state-of-the-art products can be achieved with the proposed approach. Our main conclusions and contributions can be summarized as follows:
 

\begin{itemize}
\item We propose a deep learning-based framework for large-scale HSE mapping from {medium resolution} Sentinel-2 images (10 m and 20 m GSD) with {a small amount of reference data (with a temporal gap) from Europe}. No manually labeled data is needed in the framework. 
This framework is potentially applicable for images from other satellites, such as Landsat and Sentinel-1, {and the specific network architecture {used in this study} can be replaced by other state-of-the-art architectures or improved versions}.
\item \textcolor{black}{We propose the use of a simple FCN instead of the sophisticated ones originally proposed for high resolution images, to avoid overhead and facilitate upscaling. The design choice of the framework is supported by comparisons with several baselines and investigations on the depth and width of the network as well as the experimental setup.}
\item We achieve HSE mapping results that are better than the state-of-the-art products, for several representative cities from six continents across the world. In order to carry out a fair comparison among different products and \textcolor{black}{avoid} human behavior-induced bias, two approaches for \textcolor{black}{quantitative assessment}s, in addition to city-scale and building block-scale visualizations, are performed. {Differences among HSE-related datasets are analyzed}. HSE mapping examples at regional and country scale demonstrate the general performance of the framework.
\end{itemize}
%
We hope that our work encourages the explorations of the deep-learning-based approaches along with the rich array of geo-coded products for large-scale urban mapping. \textcolor[rgb]{0,0,0}{To this end, we will publish the trained models so that researchers can extract the HSE information of a specific region of interest via the proposed framework. Trained models and sample data are available at \url{https://github.com/ChunpingQiu/Human-settlement-extent-detection-from-Sentinel-2-images-via-fully-convolutional-neural-networks-}.} Our future work includes further improving the mapping results of a specific region of interest. Additionally, the newly acquired Sentinel-2 images will allow for more timely and frequent HSE mapping and the 10- and 20-meter pixel spacing of Sentinel-2 images will allow for more detailed and accurate HSE mapping than those employing multi-spectral Landsat images with 30-meter pixel spacing. The promising results also motivate us to map more detailed multi-temporal HSE information from Sentinel-2 images in future. 

\section*{Acknowledgment}
This work is jointly supported by the China Scholarship Council (CSC), the European Research Council (ERC) under the European Union’s Horizon 2020 research and innovation program (Grant Agreement No. ERC-2016-StG-714087, Acronym: So2Sat), the Helmholtz Association under the framework of the Young Investigators Group SiPEO (VH-NG-1018, www.sipeo.bgu.tum.de), and the Bavarian Academy of Sciences and Humanities in the framework of Junges Kolleg.

\appendix

\section*{References}
\bibliographystyle{model1-num-names}
\bibliography{libBUG.bib}

\begin{thebibliography}{74}
\expandafter\ifx\csname natexlab\endcsname\relax\def\natexlab#1{#1}\fi
\providecommand{\url}[1]{\texttt{#1}}
\providecommand{\href}[2]{#2}
\providecommand{\path}[1]{#1}
\providecommand{\DOIprefix}{doi:}
\providecommand{\ArXivprefix}{arXiv:}
\providecommand{\URLprefix}{URL: }
\providecommand{\Pubmedprefix}{pmid:}
\providecommand{\doi}[1]{\href{http://dx.doi.org/#1}{\path{#1}}}
\providecommand{\Pubmed}[1]{\href{pmid:#1}{\path{#1}}}
\providecommand{\bibinfo}[2]{#2}
\ifx\xfnm\relax \def\xfnm[#1]{\unskip,\space#1}\fi
\bibitem[{uni(2018)}]{united20182018}
\bibinfo{title}{{United Nations, 2018 revision of world urbanization
  prospects}}, \bibinfo{year}{2018}.
\bibitem[{Esch et~al.(2012)Esch, Taubenb{\"{o}}ck, Roth, Heldens, Felbier,
  Schmidt, Mueller, Thiel, and Dech}]{esch2012tandem}
\bibinfo{author}{T.~Esch}, \bibinfo{author}{H.~Taubenb{\"{o}}ck},
  \bibinfo{author}{A.~Roth}, \bibinfo{author}{W.~Heldens},
  \bibinfo{author}{A.~Felbier}, \bibinfo{author}{M.~Schmidt},
  \bibinfo{author}{A.~A. Mueller}, \bibinfo{author}{M.~Thiel},
  \bibinfo{author}{S.~W. Dech},
\newblock \bibinfo{title}{{TanDEM-X mission-new perspectives for the inventory
  and monitoring of global settlement patterns}},
\newblock \bibinfo{journal}{Journal of Applied Remote Sensing}
  \bibinfo{volume}{6} (\bibinfo{year}{2012}) \bibinfo{pages}{61702}.
\bibitem[{Esch et~al.(2013)Esch, Marconcini, Felbier, Roth, Heldens, Huber,
  Schwinger, Taubenb{\"{o}}ck, M{\"{u}}ller, and Dech}]{esch2013urban}
\bibinfo{author}{T.~Esch}, \bibinfo{author}{M.~Marconcini},
  \bibinfo{author}{A.~Felbier}, \bibinfo{author}{A.~Roth},
  \bibinfo{author}{W.~Heldens}, \bibinfo{author}{M.~Huber},
  \bibinfo{author}{M.~Schwinger}, \bibinfo{author}{H.~Taubenb{\"{o}}ck},
  \bibinfo{author}{A.~M{\"{u}}ller}, \bibinfo{author}{S.~Dech},
\newblock \bibinfo{title}{{Urban footprint processor -- Fully automated
  processing chain generating settlement masks from global data of the TanDEM-X
  mission}},
\newblock \bibinfo{journal}{IEEE Geoscience and Remote Sensing Letters}
  \bibinfo{volume}{10} (\bibinfo{year}{2013}) \bibinfo{pages}{1617--1621}.
\bibitem[{Pesaresi et~al.(2016)Pesaresi, Ehrlich, Ferri, Florczyk, Freire,
  Halkia, Julea, Kemper, Soille, and Syrris}]{pesaresi2016operating}
\bibinfo{author}{M.~Pesaresi}, \bibinfo{author}{D.~Ehrlich},
  \bibinfo{author}{S.~Ferri}, \bibinfo{author}{A.~Florczyk},
  \bibinfo{author}{S.~Freire}, \bibinfo{author}{M.~Halkia},
  \bibinfo{author}{A.~Julea}, \bibinfo{author}{T.~Kemper},
  \bibinfo{author}{P.~Soille}, \bibinfo{author}{V.~Syrris},
\newblock \bibinfo{title}{{Operating procedure for the production of the Global
  Human Settlement Layer from Landsat data of the epochs 1975, 1990, 2000, and
  2014}},
\newblock \bibinfo{journal}{Publications Office of the European Union}
  (\bibinfo{year}{2016}) \bibinfo{pages}{1--62}.
\bibitem[{Corbane et~al.(2017)Corbane, Pesaresi, Politis, Syrris, Florczyk,
  Soille, Maffenini, Burger, Vasilev, Rodriguez et~al.}]{corbane2017big}
\bibinfo{author}{C.~Corbane}, \bibinfo{author}{M.~Pesaresi},
  \bibinfo{author}{P.~Politis}, \bibinfo{author}{V.~Syrris},
  \bibinfo{author}{A.~J. Florczyk}, \bibinfo{author}{P.~Soille},
  \bibinfo{author}{L.~Maffenini}, \bibinfo{author}{A.~Burger},
  \bibinfo{author}{V.~Vasilev}, \bibinfo{author}{D.~Rodriguez}, et~al.,
\newblock \bibinfo{title}{Big earth data analytics on {S}entinel-1 and
  {L}andsat imagery in support to global human settlements mapping},
\newblock \bibinfo{journal}{Big Earth Data} \bibinfo{volume}{1}
  (\bibinfo{year}{2017}) \bibinfo{pages}{118--144}.
\bibitem[{Chen et~al.(2017)Chen, Cao, Peng, and Ren}]{chen2017analysis}
\bibinfo{author}{J.~Chen}, \bibinfo{author}{X.~Cao}, \bibinfo{author}{S.~Peng},
  \bibinfo{author}{H.~Ren},
\newblock \bibinfo{title}{{Analysis and applications of GlobeLand30: a
  review}},
\newblock \bibinfo{journal}{ISPRS International Journal of Geo-Information}
  \bibinfo{volume}{6} (\bibinfo{year}{2017}) \bibinfo{pages}{230}.
\bibitem[{Wang et~al.(2017)Wang, Huang, {Brown de Colstoun}, Tilton, and
  Tan}]{Wang2017}
\bibinfo{author}{P.~Wang}, \bibinfo{author}{C.~Huang}, \bibinfo{author}{E.~C.
  {Brown de Colstoun}}, \bibinfo{author}{J.~C. Tilton},
  \bibinfo{author}{B.~Tan},
\newblock \bibinfo{title}{{Documentation for the Global Human Built-up And
  Settlement Extent (HBASE) Dataset From Landsat}},
\newblock \bibinfo{journal}{Palisades, NY: NASA Socioeconomic Data and
  Applications Center (SEDAC)}  (\bibinfo{year}{2017}).
  \bibinfo{note}{\url{https://doi.org/10.7927/H4DN434S}. Accessed 2019-04-23}.
\bibitem[{Gong et~al.(2013)Gong, Wang, Yu, Zhao, Zhao, Liang, Niu, Huang, Fu,
  Liu et~al.}]{gong2013finer}
\bibinfo{author}{P.~Gong}, \bibinfo{author}{J.~Wang}, \bibinfo{author}{L.~Yu},
  \bibinfo{author}{Y.~Zhao}, \bibinfo{author}{Y.~Zhao},
  \bibinfo{author}{L.~Liang}, \bibinfo{author}{Z.~Niu},
  \bibinfo{author}{X.~Huang}, \bibinfo{author}{H.~Fu},
  \bibinfo{author}{S.~Liu}, et~al.,
\newblock \bibinfo{title}{{Finer resolution observation and monitoring of
  global land cover: First mapping results with Landsat TM and ETM+ data}},
\newblock \bibinfo{journal}{International Journal of Remote Sensing}
  \bibinfo{volume}{34} (\bibinfo{year}{2013}) \bibinfo{pages}{2607--2654}.
\bibitem[{Gong et~al.(2019)Gong, Liu, Zhang, Li, Wang, Huang, Clinton, Ji, Li,
  Bai et~al.}]{gong2019stable}
\bibinfo{author}{P.~Gong}, \bibinfo{author}{H.~Liu},
  \bibinfo{author}{M.~Zhang}, \bibinfo{author}{C.~Li},
  \bibinfo{author}{J.~Wang}, \bibinfo{author}{H.~Huang},
  \bibinfo{author}{N.~Clinton}, \bibinfo{author}{L.~Ji},
  \bibinfo{author}{W.~Li}, \bibinfo{author}{Y.~Bai}, et~al.,
\newblock \bibinfo{title}{Stable classification with limited sample:
  transferring a 30-m resolution sample set collected in 2015 to mapping 10-m
  resolution global land cover in 2017},
\newblock \bibinfo{journal}{Science Bulletin} \bibinfo{volume}{64}
  (\bibinfo{year}{2019}) \bibinfo{pages}{370--373}.
\bibitem[{Bartholome and Belward(2005)}]{bartholome2005glc2000}
\bibinfo{author}{E.~Bartholome}, \bibinfo{author}{A.~S. Belward},
\newblock \bibinfo{title}{{GLC2000: a new approach to global land cover mapping
  from Earth observation data}},
\newblock \bibinfo{journal}{International Journal of Remote Sensing}
  \bibinfo{volume}{26} (\bibinfo{year}{2005}) \bibinfo{pages}{1959--1977}.
\bibitem[{Friedl et~al.(2002)Friedl, McIver, Hodges, Zhang, Muchoney, Strahler,
  Woodcock, Gopal, Schneider, Cooper, and Others}]{friedl2002global}
\bibinfo{author}{M.~A. Friedl}, \bibinfo{author}{D.~K. McIver},
  \bibinfo{author}{J.~C.~F. Hodges}, \bibinfo{author}{X.~Y. Zhang},
  \bibinfo{author}{D.~Muchoney}, \bibinfo{author}{A.~H. Strahler},
  \bibinfo{author}{C.~E. Woodcock}, \bibinfo{author}{S.~Gopal},
  \bibinfo{author}{A.~Schneider}, \bibinfo{author}{A.~Cooper},
  \bibinfo{author}{Others},
\newblock \bibinfo{title}{{Global land cover mapping from MODIS: algorithms and
  early results}},
\newblock \bibinfo{journal}{Remote sensing of Environment} \bibinfo{volume}{83}
  (\bibinfo{year}{2002}) \bibinfo{pages}{287--302}.
\bibitem[{Marconcini et~al.(2019)Marconcini, Metz-Marconcini, {\"U}reyen,
  Palacios-Lopez, Hanke, Bachofer, Zeidler, Esch, Gorelick, Kakarla
  et~al.}]{marconcini2019outlining}
\bibinfo{author}{M.~Marconcini}, \bibinfo{author}{A.~Metz-Marconcini},
  \bibinfo{author}{S.~{\"U}reyen}, \bibinfo{author}{D.~Palacios-Lopez},
  \bibinfo{author}{W.~Hanke}, \bibinfo{author}{F.~Bachofer},
  \bibinfo{author}{J.~Zeidler}, \bibinfo{author}{T.~Esch},
  \bibinfo{author}{N.~Gorelick}, \bibinfo{author}{A.~Kakarla}, et~al.,
\newblock \bibinfo{title}{Outlining where humans live--the world settlement
  footprint 2015},
\newblock \bibinfo{journal}{arXiv preprint arXiv:1910.12707}
  (\bibinfo{year}{2019}).
\bibitem[{Patel et~al.(2015)Patel, Angiuli, Gamba, Gaughan, Lisini, Stevens,
  Tatem, and Trianni}]{patel2015multitemporal}
\bibinfo{author}{N.~N. Patel}, \bibinfo{author}{E.~Angiuli},
  \bibinfo{author}{P.~Gamba}, \bibinfo{author}{A.~Gaughan},
  \bibinfo{author}{G.~Lisini}, \bibinfo{author}{F.~R. Stevens},
  \bibinfo{author}{A.~J. Tatem}, \bibinfo{author}{G.~Trianni},
\newblock \bibinfo{title}{{Multitemporal settlement and population mapping from
  Landsat using Google Earth Engine}},
\newblock \bibinfo{journal}{International Journal of Applied Earth Observation
  and Geoinformation} \bibinfo{volume}{35} (\bibinfo{year}{2015})
  \bibinfo{pages}{199--208}.
\bibitem[{Goldblatt et~al.(2018)Goldblatt, Stuhlmacher, Tellman, Clinton,
  Hanson, Georgescu, Wang, Serrano-Candela, Khandelwal, Cheng, and
  Others}]{goldblatt2018using}
\bibinfo{author}{R.~Goldblatt}, \bibinfo{author}{M.~F. Stuhlmacher},
  \bibinfo{author}{B.~Tellman}, \bibinfo{author}{N.~Clinton},
  \bibinfo{author}{G.~Hanson}, \bibinfo{author}{M.~Georgescu},
  \bibinfo{author}{C.~Wang}, \bibinfo{author}{F.~Serrano-Candela},
  \bibinfo{author}{A.~K. Khandelwal}, \bibinfo{author}{W.-H. Cheng},
  \bibinfo{author}{Others},
\newblock \bibinfo{title}{{Using Landsat and nighttime lights for supervised
  pixel-based image classification of urban land cover}},
\newblock \bibinfo{journal}{Remote Sensing of Environment}
  \bibinfo{volume}{205} (\bibinfo{year}{2018}) \bibinfo{pages}{253--275}.
\bibitem[{Liu et~al.(2018)Liu, Hu, Chen, Li, Xu, Li, Pei, and
  Wang}]{liu2018high}
\bibinfo{author}{X.~Liu}, \bibinfo{author}{G.~Hu}, \bibinfo{author}{Y.~Chen},
  \bibinfo{author}{X.~Li}, \bibinfo{author}{X.~Xu}, \bibinfo{author}{S.~Li},
  \bibinfo{author}{F.~Pei}, \bibinfo{author}{S.~Wang},
\newblock \bibinfo{title}{{High-resolution multi-temporal mapping of global
  urban land using Landsat images based on the Google Earth Engine Platform}},
\newblock \bibinfo{journal}{Remote Sensing of Environment}
  \bibinfo{volume}{209} (\bibinfo{year}{2018}) \bibinfo{pages}{227--239}.
\bibitem[{Xu et~al.(2018)Xu, Liu, and Xu}]{xu2018extraction}
\bibinfo{author}{R.~Xu}, \bibinfo{author}{J.~Liu}, \bibinfo{author}{J.~Xu},
\newblock \bibinfo{title}{{Extraction of high-precision urban impervious
  surfaces from Sentinel-2 multispectral imagery via modified linear spectral
  mixture analysis}},
\newblock \bibinfo{journal}{Sensors} \bibinfo{volume}{18}
  (\bibinfo{year}{2018}) \bibinfo{pages}{2873}.
\bibitem[{Qiu et~al.(2019)Qiu, Mou, Schmitt, and Zhu}]{qiuRcnn}
\bibinfo{author}{C.~Qiu}, \bibinfo{author}{L.~Mou},
  \bibinfo{author}{M.~Schmitt}, \bibinfo{author}{X.~X. Zhu},
\newblock \bibinfo{title}{{LCZ-based urban land cover classification from
  multi-seasonal Sentinel-2 images with a recurrent residual network}},
\newblock \bibinfo{journal}{ISPRS J. Photogramm. Remote Sens.}
  \bibinfo{volume}{154} (\bibinfo{year}{2019}) \bibinfo{pages}{151--162}.
\bibitem[{Ban et~al.(2015)Ban, Jacob, and Gamba}]{ban2015spaceborne}
\bibinfo{author}{Y.~Ban}, \bibinfo{author}{A.~Jacob},
  \bibinfo{author}{P.~Gamba},
\newblock \bibinfo{title}{{Spaceborne SAR data for global urban mapping at 30 m
  resolution using a robust urban extractor}},
\newblock \bibinfo{journal}{ISPRS Journal of Photogrammetry and Remote Sensing}
  \bibinfo{volume}{103} (\bibinfo{year}{2015}) \bibinfo{pages}{28--37}.
\bibitem[{Chini et~al.(2018)Chini, Pelich, Hostache, Matgen, and
  Lopez-Martinez}]{chini2018towards}
\bibinfo{author}{M.~Chini}, \bibinfo{author}{R.~Pelich},
  \bibinfo{author}{R.~Hostache}, \bibinfo{author}{P.~Matgen},
  \bibinfo{author}{C.~Lopez-Martinez},
\newblock \bibinfo{title}{{Towards a 20 m global building map from Sentinel-1
  SAR Data}},
\newblock \bibinfo{journal}{Remote Sensing} \bibinfo{volume}{10}
  (\bibinfo{year}{2018}) \bibinfo{pages}{1833}.
\bibitem[{Long et~al.(2015)Long, Shelhamer, and Darrell}]{long2015fully}
\bibinfo{author}{J.~Long}, \bibinfo{author}{E.~Shelhamer},
  \bibinfo{author}{T.~Darrell},
\newblock \bibinfo{title}{{Fully convolutional networks for semantic
  segmentation}},
\newblock in: \bibinfo{booktitle}{Proc. IEEE conference on computer vision and
  pattern recognition}, \bibinfo{address}{Boston, Massachusetts},
  \bibinfo{year}{June 8--10, 2015}, pp. \bibinfo{pages}{3431--3440}.
\bibitem[{Noh et~al.(2015)Noh, Hong, and Han}]{noh2015learning}
\bibinfo{author}{H.~Noh}, \bibinfo{author}{S.~Hong}, \bibinfo{author}{B.~Han},
\newblock \bibinfo{title}{{Learning deconvolution network for semantic
  segmentation}},
\newblock in: \bibinfo{booktitle}{Proc. IEEE international conference on
  computer vision, Washington, DC, USA, 7--13 December}, \bibinfo{year}{2015},
  pp. \bibinfo{pages}{1520--1528}.
\bibitem[{Badrinarayanan et~al.(2017)Badrinarayanan, Kendall, and
  Cipolla}]{badrinarayanan2017segnet}
\bibinfo{author}{V.~Badrinarayanan}, \bibinfo{author}{A.~Kendall},
  \bibinfo{author}{R.~Cipolla},
\newblock \bibinfo{title}{{Segnet: A deep convolutional encoder-decoder
  architecture for image segmentation}},
\newblock \bibinfo{journal}{IEEE transactions on pattern analysis and machine
  intelligence} \bibinfo{volume}{39} (\bibinfo{year}{2017})
  \bibinfo{pages}{2481--2495}.
\bibitem[{Ronneberger et~al.(2015)Ronneberger, Fischer, and
  Brox}]{ronneberger2015u}
\bibinfo{author}{O.~Ronneberger}, \bibinfo{author}{P.~Fischer},
  \bibinfo{author}{T.~Brox},
\newblock \bibinfo{title}{{U-net: Convolutional networks for biomedical image
  segmentation}},
\newblock in: \bibinfo{booktitle}{International Conference on Medical image
  computing and computer-assisted intervention},
  \bibinfo{organization}{Springer}, \bibinfo{address}{Munich, Germany, 5-9
  October}, \bibinfo{year}{2015}, pp. \bibinfo{pages}{234--241}.
\bibitem[{Paisitkriangkrai et~al.(2016)Paisitkriangkrai, Sherrah, Janney, and
  Van Den~Hengel}]{paisitkriangkrai2016semantic}
\bibinfo{author}{S.~Paisitkriangkrai}, \bibinfo{author}{J.~Sherrah},
  \bibinfo{author}{P.~Janney}, \bibinfo{author}{A.~Van Den~Hengel},
\newblock \bibinfo{title}{Semantic labeling of aerial and satellite imagery},
\newblock \bibinfo{journal}{IEEE Journal of Selected Topics in Applied Earth
  Observations and Remote Sensing} \bibinfo{volume}{9} (\bibinfo{year}{2016})
  \bibinfo{pages}{2868--2881}.
\bibitem[{Maggiori et~al.(2016)Maggiori, Tarabalka, Charpiat, and
  Alliez}]{maggiori2016fully}
\bibinfo{author}{E.~Maggiori}, \bibinfo{author}{Y.~Tarabalka},
  \bibinfo{author}{G.~Charpiat}, \bibinfo{author}{P.~Alliez},
\newblock \bibinfo{title}{Fully convolutional neural networks for remote
  sensing image classification},
\newblock in: \bibinfo{booktitle}{2016 IEEE international geoscience and remote
  sensing symposium (IGARSS)}, \bibinfo{organization}{IEEE},
  \bibinfo{year}{2016}, pp. \bibinfo{pages}{5071--5074}.
\bibitem[{L{\"a}ngkvist et~al.(2016)L{\"a}ngkvist, Kiselev, Alirezaie, and
  Loutfi}]{langkvist2016classification}
\bibinfo{author}{M.~L{\"a}ngkvist}, \bibinfo{author}{A.~Kiselev},
  \bibinfo{author}{M.~Alirezaie}, \bibinfo{author}{A.~Loutfi},
\newblock \bibinfo{title}{Classification and segmentation of satellite
  orthoimagery using convolutional neural networks},
\newblock \bibinfo{journal}{Remote Sensing} \bibinfo{volume}{8}
  (\bibinfo{year}{2016}) \bibinfo{pages}{329}.
\bibitem[{Maggiori et~al.(2016)Maggiori, Tarabalka, Charpiat, and
  Alliez}]{maggiori2016convolutional}
\bibinfo{author}{E.~Maggiori}, \bibinfo{author}{Y.~Tarabalka},
  \bibinfo{author}{G.~Charpiat}, \bibinfo{author}{P.~Alliez},
\newblock \bibinfo{title}{Convolutional neural networks for large-scale
  remote-sensing image classification},
\newblock \bibinfo{journal}{IEEE Transactions on Geoscience and Remote Sensing}
  \bibinfo{volume}{55} (\bibinfo{year}{2016}) \bibinfo{pages}{645--657}.
\bibitem[{Fu et~al.(2017)Fu, Liu, Zhou, Sun, and Zhang}]{fu2017classification}
\bibinfo{author}{G.~Fu}, \bibinfo{author}{C.~Liu}, \bibinfo{author}{R.~Zhou},
  \bibinfo{author}{T.~Sun}, \bibinfo{author}{Q.~Zhang},
\newblock \bibinfo{title}{Classification for high resolution remote sensing
  imagery using a fully convolutional network},
\newblock \bibinfo{journal}{Remote Sensing} \bibinfo{volume}{9}
  (\bibinfo{year}{2017}) \bibinfo{pages}{498}.
\bibitem[{Volpi and Tuia(2016)}]{volpi2016dense}
\bibinfo{author}{M.~Volpi}, \bibinfo{author}{D.~Tuia},
\newblock \bibinfo{title}{Dense semantic labeling of subdecimeter resolution
  images with convolutional neural networks},
\newblock \bibinfo{journal}{IEEE Transactions on Geoscience and Remote Sensing}
  \bibinfo{volume}{55} (\bibinfo{year}{2016}) \bibinfo{pages}{881--893}.
\bibitem[{Ru{\ss}wurm and K{\"o}rner(2018)}]{russwurm2018multi}
\bibinfo{author}{M.~Ru{\ss}wurm}, \bibinfo{author}{M.~K{\"o}rner},
\newblock \bibinfo{title}{Multi-temporal land cover classification with
  sequential recurrent encoders},
\newblock \bibinfo{journal}{ISPRS International Journal of Geo-Information}
  \bibinfo{volume}{7} (\bibinfo{year}{2018}) \bibinfo{pages}{129}.
\bibitem[{Zhang et~al.(2019)Zhang, Sargent, Pan, Li, Gardiner, Hare, and
  Atkinson}]{zhang2019joint}
\bibinfo{author}{C.~Zhang}, \bibinfo{author}{I.~Sargent},
  \bibinfo{author}{X.~Pan}, \bibinfo{author}{H.~Li},
  \bibinfo{author}{A.~Gardiner}, \bibinfo{author}{J.~Hare},
  \bibinfo{author}{P.~M. Atkinson},
\newblock \bibinfo{title}{Joint deep learning for land cover and land use
  classification},
\newblock \bibinfo{journal}{Remote sensing of environment}
  \bibinfo{volume}{221} (\bibinfo{year}{2019}) \bibinfo{pages}{173--187}.
\bibitem[{Zhong et~al.(2019)Zhong, Hu, and Zhou}]{zhong2019deep}
\bibinfo{author}{L.~Zhong}, \bibinfo{author}{L.~Hu}, \bibinfo{author}{H.~Zhou},
\newblock \bibinfo{title}{Deep learning based multi-temporal crop
  classification},
\newblock \bibinfo{journal}{Remote sensing of environment}
  \bibinfo{volume}{221} (\bibinfo{year}{2019}) \bibinfo{pages}{430--443}.
\bibitem[{Hu et~al.(2019)Hu, Patel, Robert, Novosad, Asher, Tang, Burke,
  Lobell, and Ermon}]{hu2019mapping}
\bibinfo{author}{W.~Hu}, \bibinfo{author}{J.~H. Patel}, \bibinfo{author}{Z.-A.
  Robert}, \bibinfo{author}{P.~Novosad}, \bibinfo{author}{S.~Asher},
  \bibinfo{author}{Z.~Tang}, \bibinfo{author}{M.~Burke},
  \bibinfo{author}{D.~Lobell}, \bibinfo{author}{S.~Ermon},
\newblock \bibinfo{title}{Mapping missing population in rural india: A deep
  learning approach with satellite imagery},
\newblock \bibinfo{journal}{arXiv preprint arXiv:1905.02196}
  (\bibinfo{year}{2019}).
\bibitem[{Lang et~al.(2019)Lang, Schindler, and Wegner}]{lang2019country}
\bibinfo{author}{N.~Lang}, \bibinfo{author}{K.~Schindler},
  \bibinfo{author}{J.~D. Wegner},
\newblock \bibinfo{title}{Country-wide high-resolution vegetation height
  mapping with sentinel-2},
\newblock \bibinfo{journal}{arXiv preprint arXiv:1904.13270}
  (\bibinfo{year}{2019}).
\bibitem[{He et~al.(2018)He, Liu, Gou, Zhang, Zhang, and Xu}]{he2018detecting}
\bibinfo{author}{C.~He}, \bibinfo{author}{Z.~Liu}, \bibinfo{author}{S.~Gou},
  \bibinfo{author}{Q.~Zhang}, \bibinfo{author}{J.~Zhang},
  \bibinfo{author}{L.~Xu},
\newblock \bibinfo{title}{Detecting global urban expansion over the last three
  decades using a fully convolutional network},
\newblock \bibinfo{journal}{Environmental Research Letters}
  (\bibinfo{year}{2018}).
\bibitem[{Helber et~al.(2019)Helber, Bischke, Dengel, and
  Borth}]{helber2019eurosat}
\bibinfo{author}{P.~Helber}, \bibinfo{author}{B.~Bischke},
  \bibinfo{author}{A.~Dengel}, \bibinfo{author}{D.~Borth},
\newblock \bibinfo{title}{Eurosat: A novel dataset and deep learning benchmark
  for land use and land cover classification},
\newblock \bibinfo{journal}{IEEE Journal of Selected Topics in Applied Earth
  Observations and Remote Sensing} \bibinfo{volume}{12} (\bibinfo{year}{2019})
  \bibinfo{pages}{2217--2226}.
\bibitem[{Sumbul et~al.(2019)Sumbul, Charfuelan, Demir, and
  Markl}]{sumbul2019bigearthnet}
\bibinfo{author}{G.~Sumbul}, \bibinfo{author}{M.~Charfuelan},
  \bibinfo{author}{B.~Demir}, \bibinfo{author}{V.~Markl},
\newblock \bibinfo{title}{{BigEarthNet: A Large-Scale Benchmark Archive For
  Remote Sensing Image Understanding}},
\newblock \bibinfo{journal}{arXiv preprint arXiv:1902.06148}
  (\bibinfo{year}{2019}).
\bibitem[{Schmitt et~al.(2019)Schmitt, Hughes, Qiu, and
  Zhu}]{schmitt2019sen12ms}
\bibinfo{author}{M.~Schmitt}, \bibinfo{author}{L.~H. Hughes},
  \bibinfo{author}{C.~Qiu}, \bibinfo{author}{X.~X. Zhu},
\newblock \bibinfo{title}{{SEN12MS--A Curated Dataset of Georeferenced
  Multi-Spectral Sentinel-1/2 Imagery for Deep Learning and Data Fusion}},
\newblock \bibinfo{journal}{arXiv preprint arXiv:1906.07789}
  (\bibinfo{year}{2019}).
\bibitem[{Zhu et~al.(2017)Zhu, Tuia, Mou, Xia, Zhang, Xu, and
  Fraundorfer}]{zhu2017deep}
\bibinfo{author}{X.~X. Zhu}, \bibinfo{author}{D.~Tuia},
  \bibinfo{author}{L.~Mou}, \bibinfo{author}{G.-S. Xia},
  \bibinfo{author}{L.~Zhang}, \bibinfo{author}{F.~Xu},
  \bibinfo{author}{F.~Fraundorfer},
\newblock \bibinfo{title}{Deep learning in remote sensing: A comprehensive
  review and list of resources},
\newblock \bibinfo{journal}{IEEE Geoscience and Remote Sensing Magazine}
  \bibinfo{volume}{5} (\bibinfo{year}{2017}) \bibinfo{pages}{8--36}.
\bibitem[{Gei{\ss} et~al.(2017)Gei{\ss}, Pelizari, Schrade, Brenning, and
  Taubenb{\"o}ck}]{geiss2017effect}
\bibinfo{author}{C.~Gei{\ss}}, \bibinfo{author}{P.~A. Pelizari},
  \bibinfo{author}{H.~Schrade}, \bibinfo{author}{A.~Brenning},
  \bibinfo{author}{H.~Taubenb{\"o}ck},
\newblock \bibinfo{title}{On the effect of spatially non-disjoint training and
  test samples on estimated model generalization capabilities in supervised
  classification with spatial features},
\newblock \bibinfo{journal}{IEEE Geoscience and Remote Sensing Letters}
  \bibinfo{volume}{14} (\bibinfo{year}{2017}) \bibinfo{pages}{2008--2012}.
\bibitem[{Drusch et~al.(2012)Drusch, Del~Bello, Carlier, Colin, Fernandez,
  Gascon, Hoersch, Isola, Laberinti, Martimort et~al.}]{drusch2012sentinel}
\bibinfo{author}{M.~Drusch}, \bibinfo{author}{U.~Del~Bello},
  \bibinfo{author}{S.~Carlier}, \bibinfo{author}{O.~Colin},
  \bibinfo{author}{V.~Fernandez}, \bibinfo{author}{F.~Gascon},
  \bibinfo{author}{B.~Hoersch}, \bibinfo{author}{C.~Isola},
  \bibinfo{author}{P.~Laberinti}, \bibinfo{author}{P.~Martimort}, et~al.,
\newblock \bibinfo{title}{Sentinel-2: Esa's optical high-resolution mission for
  gmes operational services},
\newblock \bibinfo{journal}{Remote sensing of Environment}
  \bibinfo{volume}{120} (\bibinfo{year}{2012}) \bibinfo{pages}{25--36}.
\bibitem[{Gorelick et~al.(2017)Gorelick, Hancher, Dixon, Ilyushchenko, Thau,
  and Moore}]{gorelick2017google}
\bibinfo{author}{N.~Gorelick}, \bibinfo{author}{M.~Hancher},
  \bibinfo{author}{M.~Dixon}, \bibinfo{author}{S.~Ilyushchenko},
  \bibinfo{author}{D.~Thau}, \bibinfo{author}{R.~Moore},
\newblock \bibinfo{title}{{Google Earth Engine: Planetary-scale geospatial
  analysis for everyone}},
\newblock \bibinfo{journal}{Remote Sensing of Environment}
  \bibinfo{volume}{202} (\bibinfo{year}{2017}) \bibinfo{pages}{18--27}.
\bibitem[{Schmitt et~al.(2019)Schmitt, Hughes, Qiu, and Zhu}]{Aggregating}
\bibinfo{author}{M.~Schmitt}, \bibinfo{author}{L.~H. Hughes},
  \bibinfo{author}{C.~Qiu}, \bibinfo{author}{X.~X. Zhu},
\newblock \bibinfo{title}{{Aggregating Cloud-Free Sentinel-2 Images with Google
  Earth Engine}},
\newblock in: \bibinfo{booktitle}{MRSS19 - Munich Remote Sensing Symposium
  2019}, \bibinfo{year}{2019}.
\bibitem[{Langanke(2016)}]{CopernicusHRLI}
\bibinfo{author}{T.~Langanke},
\newblock \bibinfo{title}{{Copernicus Land Monitoring Service – High
  Resolution Layer Imperviousness: Product Specifications Document}},
\newblock \bibinfo{journal}{Copernicus team at EEA}  (\bibinfo{year}{2016}).
\bibitem[{He et~al.(2016)He, Zhang, Ren, and Sun}]{he2016deep}
\bibinfo{author}{K.~He}, \bibinfo{author}{X.~Zhang}, \bibinfo{author}{S.~Ren},
  \bibinfo{author}{J.~Sun},
\newblock \bibinfo{title}{Deep residual learning for image recognition},
\newblock in: \bibinfo{booktitle}{Proceedings of the IEEE conference on
  computer vision and pattern recognition}, \bibinfo{year}{2016}, pp.
  \bibinfo{pages}{770--778}.
\bibitem[{Xie et~al.(2017)Xie, Girshick, Doll{\'a}r, Tu, and
  He}]{xie2017aggregated}
\bibinfo{author}{S.~Xie}, \bibinfo{author}{R.~Girshick},
  \bibinfo{author}{P.~Doll{\'a}r}, \bibinfo{author}{Z.~Tu},
  \bibinfo{author}{K.~He},
\newblock \bibinfo{title}{Aggregated residual transformations for deep neural
  networks},
\newblock in: \bibinfo{booktitle}{CVPR}, \bibinfo{year}{2017}, pp.
  \bibinfo{pages}{1492--1500}.
\bibitem[{Szegedy et~al.(2015)Szegedy, Liu, Jia, Sermanet, Reed, Anguelov,
  Erhan, Vanhoucke, and Rabinovich}]{szegedy2015going}
\bibinfo{author}{C.~Szegedy}, \bibinfo{author}{W.~Liu},
  \bibinfo{author}{Y.~Jia}, \bibinfo{author}{P.~Sermanet},
  \bibinfo{author}{S.~Reed}, \bibinfo{author}{D.~Anguelov},
  \bibinfo{author}{D.~Erhan}, \bibinfo{author}{V.~Vanhoucke},
  \bibinfo{author}{A.~Rabinovich},
\newblock \bibinfo{title}{Going deeper with convolutions},
\newblock in: \bibinfo{booktitle}{Proceedings of the IEEE conference on
  computer vision and pattern recognition}, \bibinfo{year}{2015}, pp.
  \bibinfo{pages}{1--9}.
\bibitem[{Chollet(2017)}]{chollet17xception}
\bibinfo{author}{F.~Chollet},
\newblock \bibinfo{title}{Xception: Deep learning with depthwise separable
  convolutions},
\newblock in: \bibinfo{booktitle}{The IEEE Conference on Computer Vision and
  Pattern Recognition (CVPR)}, \bibinfo{year}{2017}.
\bibitem[{Hua et~al.(2019{\natexlab{a}})Hua, Mou, and Zhu}]{hua2019relation}
\bibinfo{author}{Y.~Hua}, \bibinfo{author}{L.~Mou}, \bibinfo{author}{X.~X.
  Zhu},
\newblock \bibinfo{title}{Relation network for multi-label aerial image
  classification},
\newblock \bibinfo{journal}{arXiv:1907.07274}
  (\bibinfo{year}{2019}{\natexlab{a}}).
\bibitem[{Hua et~al.(2019{\natexlab{b}})Hua, Mou, and Zhu}]{hua2019recurrently}
\bibinfo{author}{Y.~Hua}, \bibinfo{author}{L.~Mou}, \bibinfo{author}{X.~X.
  Zhu},
\newblock \bibinfo{title}{Recurrently exploring class-wise attention in a
  hybrid convolutional and bidirectional {LSTM} network for multi-label aerial
  image classification},
\newblock \bibinfo{journal}{ISPRS Journal of Photogrammetry and Remote Sensing}
  \bibinfo{volume}{149} (\bibinfo{year}{2019}{\natexlab{b}})
  \bibinfo{pages}{188--199}.
\bibitem[{Zhu et~al.(2019)Zhu, Hu, Qiu, Shi, Kang, Mou, Bagheri, H{\"a}berle,
  Hua, Huang et~al.}]{zhu2019so2sat}
\bibinfo{author}{X.~X. Zhu}, \bibinfo{author}{J.~Hu}, \bibinfo{author}{C.~Qiu},
  \bibinfo{author}{Y.~Shi}, \bibinfo{author}{J.~Kang},
  \bibinfo{author}{L.~Mou}, \bibinfo{author}{H.~Bagheri},
  \bibinfo{author}{M.~H{\"a}berle}, \bibinfo{author}{Y.~Hua},
  \bibinfo{author}{R.~Huang}, et~al.,
\newblock \bibinfo{title}{{So2Sat LCZ42: A benchmark dataset for global local
  climate zones classification}},
\newblock \bibinfo{journal}{arXiv preprint arXiv:1912.12171}
  (\bibinfo{year}{2019}).
\bibitem[{Hasanpour et~al.(2016)Hasanpour, Rouhani, Fayyaz, and
  Sabokrou}]{hasanpour2016lets}
\bibinfo{author}{S.~H. Hasanpour}, \bibinfo{author}{M.~Rouhani},
  \bibinfo{author}{M.~Fayyaz}, \bibinfo{author}{M.~Sabokrou},
\newblock \bibinfo{title}{Lets keep it simple, using simple architectures to
  outperform deeper and more complex architectures},
\newblock \bibinfo{journal}{arXiv preprint arXiv:1608.06037}
  (\bibinfo{year}{2016}).
\bibitem[{He et~al.(2015)He, Zhang, Ren, and Sun}]{he2015delving}
\bibinfo{author}{K.~He}, \bibinfo{author}{X.~Zhang}, \bibinfo{author}{S.~Ren},
  \bibinfo{author}{J.~Sun},
\newblock \bibinfo{title}{Delving deep into rectifiers: Surpassing human-level
  performance on imagenet classification},
\newblock in: \bibinfo{booktitle}{Proceedings of the IEEE international
  conference on computer vision}, \bibinfo{year}{2015}, pp.
  \bibinfo{pages}{1026--1034}.
\bibitem[{Chollet et~al.(2015)}]{chollet2015keras}
\bibinfo{author}{F.~Chollet}, et~al., \bibinfo{title}{Keras},
  \bibinfo{howpublished}{\url{https://keras.io}}, \bibinfo{year}{2015}.
\bibitem[{Schneider et~al.(2010)Schneider, Friedl, and Potere}]{Schneider.2010}
\bibinfo{author}{A.~Schneider}, \bibinfo{author}{M.~A. Friedl},
  \bibinfo{author}{D.~Potere},
\newblock \bibinfo{title}{{Mapping global urban areas using MODIS 500-m data:
  New methods and datasets based on `urban ecoregions'}},
\newblock \bibinfo{journal}{Remote Sensing of Environment}
  \bibinfo{volume}{114} (\bibinfo{year}{2010}) \bibinfo{pages}{1733--1746}.
\bibitem[{Fan et~al.(2014)Fan, Zipf, Fu, and Neis}]{fan2014quality}
\bibinfo{author}{H.~Fan}, \bibinfo{author}{A.~Zipf}, \bibinfo{author}{Q.~Fu},
  \bibinfo{author}{P.~Neis},
\newblock \bibinfo{title}{{Quality assessment for building footprints data on
  OpenStreetMap}},
\newblock \bibinfo{journal}{International Journal of Geographical Information
  Science} \bibinfo{volume}{28} (\bibinfo{year}{2014})
  \bibinfo{pages}{700--719}.
\bibitem[{Arsanjani et~al.(2015)Arsanjani, Mooney, Zipf, and
  Schauss}]{arsanjani2015quality}
\bibinfo{author}{J.~J. Arsanjani}, \bibinfo{author}{P.~Mooney},
  \bibinfo{author}{A.~Zipf}, \bibinfo{author}{A.~Schauss},
\newblock \bibinfo{title}{{Quality assessment of the contributed land use
  information from OpenStreetMap versus authoritative datasets}},
\newblock in: \bibinfo{booktitle}{OpenStreetMap in GIScience},
  \bibinfo{year}{2015}, pp. \bibinfo{pages}{37--58}.
\bibitem[{Johnson et~al.(2017)Johnson, Iizuka, Bragais, Endo, and
  Magcale-Macandog}]{johnson2017employing}
\bibinfo{author}{B.~A. Johnson}, \bibinfo{author}{K.~Iizuka},
  \bibinfo{author}{M.~A. Bragais}, \bibinfo{author}{I.~Endo},
  \bibinfo{author}{D.~B. Magcale-Macandog},
\newblock \bibinfo{title}{{Employing crowdsourced geographic data and
  multi-temporal/multi-sensor satellite imagery to monitor land cover change: A
  case study in an urbanizing region of the Philippines}},
\newblock \bibinfo{journal}{Computers, Environment and Urban Systems}
  \bibinfo{volume}{64} (\bibinfo{year}{2017}) \bibinfo{pages}{184--193}.
\bibitem[{Viana et~al.(2019)Viana, Encalada, and Rocha}]{viana2019value}
\bibinfo{author}{C.~M. Viana}, \bibinfo{author}{L.~Encalada},
  \bibinfo{author}{J.~Rocha},
\newblock \bibinfo{title}{{The value of OpenStreetMap Historical Contributions
  as a Source of Sampling Data for Multi-temporal Land Use/Cover Maps}},
\newblock \bibinfo{journal}{ISPRS International Journal of Geo-Information}
  \bibinfo{volume}{8} (\bibinfo{year}{2019}) \bibinfo{pages}{116}.
\bibitem[{Zhao et~al.(2017)Zhao, Shi, Qi, Wang, and Jia}]{zhao2017pyramid}
\bibinfo{author}{H.~Zhao}, \bibinfo{author}{J.~Shi}, \bibinfo{author}{X.~Qi},
  \bibinfo{author}{X.~Wang}, \bibinfo{author}{J.~Jia},
\newblock \bibinfo{title}{Pyramid scene parsing network},
\newblock in: \bibinfo{booktitle}{Proceedings of the IEEE conference on
  computer vision and pattern recognition}, \bibinfo{year}{2017}, pp.
  \bibinfo{pages}{2881--2890}.
\bibitem[{Fu et~al.(2018)Fu, Liu, Tian, Fang, and Lu}]{fu2018dual}
\bibinfo{author}{J.~Fu}, \bibinfo{author}{J.~Liu}, \bibinfo{author}{H.~Tian},
  \bibinfo{author}{Z.~Fang}, \bibinfo{author}{H.~Lu},
\newblock \bibinfo{title}{{Dual attention network for scene segmentation}},
\newblock \bibinfo{journal}{arXiv preprint arXiv:1809.02983}
  (\bibinfo{year}{2018}).
\bibitem[{Esch et~al.(2017)Esch, Heldens, Hirner, Keil, Marconcini, Roth,
  Zeidler, Dech, and Strano}]{esch2017breaking}
\bibinfo{author}{T.~Esch}, \bibinfo{author}{W.~Heldens},
  \bibinfo{author}{A.~Hirner}, \bibinfo{author}{M.~Keil},
  \bibinfo{author}{M.~Marconcini}, \bibinfo{author}{A.~Roth},
  \bibinfo{author}{J.~Zeidler}, \bibinfo{author}{S.~Dech},
  \bibinfo{author}{E.~Strano},
\newblock \bibinfo{title}{{Breaking new ground in mapping human settlements
  from space--The Global Urban Footprint}},
\newblock \bibinfo{journal}{ISPRS Journal of Photogrammetry and Remote Sensing}
  \bibinfo{volume}{134} (\bibinfo{year}{2017}) \bibinfo{pages}{30--42}.
\bibitem[{Klotz et~al.(2016)Klotz, Kemper, Gei{\ss}, Esch, and
  Taubenb{\"o}ck}]{klotz2016good}
\bibinfo{author}{M.~Klotz}, \bibinfo{author}{T.~Kemper},
  \bibinfo{author}{C.~Gei{\ss}}, \bibinfo{author}{T.~Esch},
  \bibinfo{author}{H.~Taubenb{\"o}ck},
\newblock \bibinfo{title}{How good is the map? a multi-scale cross-comparison
  framework for global settlement layers: Evidence from central europe},
\newblock \bibinfo{journal}{Remote Sensing of Environment}
  \bibinfo{volume}{178} (\bibinfo{year}{2016}) \bibinfo{pages}{191--212}.
\bibitem[{Qiu et~al.(2018)Qiu, Schmitt, and Zhu}]{qiu2018towards}
\bibinfo{author}{C.~Qiu}, \bibinfo{author}{M.~Schmitt}, \bibinfo{author}{X.~X.
  Zhu},
\newblock \bibinfo{title}{{Towards automatic SAR-optical stereogrammetry over
  urban areas using very high resolution imagery}},
\newblock \bibinfo{journal}{ISPRS Journal of Photogrammetry and Remote Sensing}
  \bibinfo{volume}{138} (\bibinfo{year}{2018}) \bibinfo{pages}{218--231}.
\bibitem[{Stengel et~al.(2017)Stengel, Stapelberg, Sus, Schlundt, Poulsen,
  Thomas, Christensen, {Carbajal Henken}, Preusker, Fischer, and
  Others}]{stengel2017cloud}
\bibinfo{author}{M.~Stengel}, \bibinfo{author}{S.~Stapelberg},
  \bibinfo{author}{O.~Sus}, \bibinfo{author}{C.~Schlundt},
  \bibinfo{author}{C.~Poulsen}, \bibinfo{author}{G.~Thomas},
  \bibinfo{author}{M.~Christensen}, \bibinfo{author}{C.~{Carbajal Henken}},
  \bibinfo{author}{R.~Preusker}, \bibinfo{author}{J.~Fischer},
  \bibinfo{author}{Others},
\newblock \bibinfo{title}{{Cloud property datasets retrieved from AVHRR, MODIS,
  AATSR and MERIS in the framework of the Cloud{\_}cci project}},
\newblock \bibinfo{journal}{Earth System Science Data} \bibinfo{volume}{9}
  (\bibinfo{year}{2017}) \bibinfo{pages}{881--904}.
\bibitem[{Schmitt and Zhu(2016)}]{schmitt2016data}
\bibinfo{author}{M.~Schmitt}, \bibinfo{author}{X.~X. Zhu},
\newblock \bibinfo{title}{{Data fusion and remote sensing: An ever-growing
  relationship}},
\newblock \bibinfo{journal}{IEEE Geoscience and Remote Sensing Magazine}
  \bibinfo{volume}{4} (\bibinfo{year}{2016}) \bibinfo{pages}{6--23}.
\bibitem[{Ghamisi et~al.(2018)Ghamisi, Rasti, Yokoya, Wang, Hofle, Bruzzone,
  Bovolo, Chi, Anders, Gloaguen, and Others}]{ghamisi2018multisource}
\bibinfo{author}{P.~Ghamisi}, \bibinfo{author}{B.~Rasti},
  \bibinfo{author}{N.~Yokoya}, \bibinfo{author}{Q.~Wang},
  \bibinfo{author}{B.~Hofle}, \bibinfo{author}{L.~Bruzzone},
  \bibinfo{author}{F.~Bovolo}, \bibinfo{author}{M.~Chi},
  \bibinfo{author}{K.~Anders}, \bibinfo{author}{R.~Gloaguen},
  \bibinfo{author}{Others},
\newblock \bibinfo{title}{{Multisource and multitemporal data fusion in remote
  sensing}},
\newblock \bibinfo{journal}{arXiv preprint arXiv:1812.08287}
  (\bibinfo{year}{2018}).
\bibitem[{Lefebvre et~al.(2016)Lefebvre, Sannier, and
  Corpetti}]{lefebvre2016monitoring}
\bibinfo{author}{A.~Lefebvre}, \bibinfo{author}{C.~Sannier},
  \bibinfo{author}{T.~Corpetti},
\newblock \bibinfo{title}{{Monitoring urban areas with Sentinel-2A data:
  Application to the update of the Copernicus high resolution layer
  imperviousness degree}},
\newblock \bibinfo{journal}{Remote Sensing} \bibinfo{volume}{8}
  (\bibinfo{year}{2016}) \bibinfo{pages}{606}.
\bibitem[{Hong et~al.(2019{\natexlab{a}})Hong, Yokoya, Chanussot, and
  Zhu}]{hong2019cospace}
\bibinfo{author}{D.~Hong}, \bibinfo{author}{N.~Yokoya},
  \bibinfo{author}{J.~Chanussot}, \bibinfo{author}{X.~X. Zhu},
\newblock \bibinfo{title}{Co{S}pace: Common subspace learning from
  hyperspectral-multispectral correspondences},
\newblock \bibinfo{journal}{IEEE Transactions on Geoscience and Remote Sensing}
  \bibinfo{volume}{57} (\bibinfo{year}{2019}{\natexlab{a}})
  \bibinfo{pages}{4349--4359}.
\bibitem[{Hong et~al.(2019{\natexlab{b}})Hong, Yokoya, Ge, Chanussot, and
  Zhu}]{hong2019learnable}
\bibinfo{author}{D.~Hong}, \bibinfo{author}{N.~Yokoya},
  \bibinfo{author}{N.~Ge}, \bibinfo{author}{J.~Chanussot},
  \bibinfo{author}{X.~X. Zhu},
\newblock \bibinfo{title}{Learnable manifold alignment ({L}e{MA}): A
  semi-supervised cross-modality learning framework for land cover and land use
  classification},
\newblock \bibinfo{journal}{ISPRS Journal of Photogrammetry and Remote Sensing}
  \bibinfo{volume}{147} (\bibinfo{year}{2019}{\natexlab{b}})
  \bibinfo{pages}{193--205}.
\bibitem[{Hong et~al.(2019{\natexlab{c}})Hong, Yokoya, Chanussot, and
  Zhu}]{hong2019augmented}
\bibinfo{author}{D.~Hong}, \bibinfo{author}{N.~Yokoya},
  \bibinfo{author}{J.~Chanussot}, \bibinfo{author}{X.~X. Zhu},
\newblock \bibinfo{title}{An augmented linear mixing model to address spectral
  variability for hyperspectral unmixing},
\newblock \bibinfo{journal}{IEEE Transactions on Image Processing}
  \bibinfo{volume}{28} (\bibinfo{year}{2019}{\natexlab{c}})
  \bibinfo{pages}{1923--1938}.
\bibitem[{Qiu et~al.(2019)Qiu, Mou, Schmitt, and Zhu}]{qiu2019fusingLetter}
\bibinfo{author}{C.~Qiu}, \bibinfo{author}{L.~Mou},
  \bibinfo{author}{M.~Schmitt}, \bibinfo{author}{X.~X. Zhu},
\newblock \bibinfo{title}{Fusing multi-seasonal sentinel-2 imagery for urban
  land cover classification with residual convolutional neural networks}
  (\bibinfo{year}{2019}). \bibinfo{note}{DOI: 10.1109/LGRS.2019.2953497}.
\bibitem[{Maggiolo et~al.(2018)Maggiolo, Marcos, Moser, and
  Tuia}]{maggiolo2018improving}
\bibinfo{author}{L.~Maggiolo}, \bibinfo{author}{D.~Marcos},
  \bibinfo{author}{G.~Moser}, \bibinfo{author}{D.~Tuia},
\newblock \bibinfo{title}{{Improving Maps from CNNs Trained with Sparse,
  Scribbled Ground Truths Using Fully Connected CRFs}},
\newblock in: \bibinfo{booktitle}{IGARSS 2018-2018 IEEE International
  Geoscience and Remote Sensing Symposium}, \bibinfo{organization}{IEEE},
  \bibinfo{year}{2018}, pp. \bibinfo{pages}{2099--2102}.
\bibitem[{Tuia et~al.(2016)Tuia, Persello, and Bruzzone}]{tuia2016domain}
\bibinfo{author}{D.~Tuia}, \bibinfo{author}{C.~Persello},
  \bibinfo{author}{L.~Bruzzone},
\newblock \bibinfo{title}{{Domain adaptation for the classification of remote
  sensing data: An overview of recent advances}},
\newblock \bibinfo{journal}{IEEE geoscience and Remote Sensing magazine}
  \bibinfo{volume}{4} (\bibinfo{year}{2016}) \bibinfo{pages}{41--57}.

\end{thebibliography}

\end{document}